\documentclass[acmtog]{acmart}
\acmSubmissionID{1134}

\AtBeginDocument{%
  }

\usepackage{makecell}

\usepackage{multirow}
\usepackage{subcaption}
\usepackage{colortbl}
\usepackage{cleveref}

\usepackage{bm}

\newcommand{\bu}{\bm{u}}
\newcommand{\bo}{\bm{\omega_o}}
\newcommand{\bi}{\bm{\omega_i}}

\newcommand{\toff}{T_{\rm off}}
\newcommand{\trgb}{T_{\rm rgb}}
\newcommand{\moff}{M_{\rm off}}
\newcommand{\mrgb}{M_{\rm rgb}}

\copyrightyear{2026}
\acmYear{2026}
\setcopyright{cc}
\setcctype{by}
\acmConference[SIGGRAPH Conference Papers '26]{Special Interest Group on Computer Graphics and Interactive Techniques Conference Conference Papers}{July 19--23, 2026}{Los Angeles, CA, USA}
\acmBooktitle{Special Interest Group on Computer Graphics and Interactive Techniques Conference Conference Papers (SIGGRAPH Conference Papers '26), July 19--23, 2026, Los Angeles, CA, USA}
\acmDOI{10.1145/3799902.3811190}
\acmISBN{979-8-4007-2554-8/2026/07}

\begin{document}

\title{VideoNeuMat: Neural Material Extraction from Generative Video Models}


\author{Bowen Xue}
\authornote{Equal contribution.}
\email{bowen.xue@manchester.ac.uk}
\orcid{0000-0002-6628-577X}
\affiliation{%
  \institution{University of Manchester}
  \city{Manchester}
  \country{United Kingdom}
}

\author{Saeed Hadadan}
\authornotemark[1]
\email{shadadan@nvidia.com}
\orcid{0000-0003-1614-9531}
\affiliation{%
  \institution{NVIDIA}
  \city{Santa Clara}
  \country{United States of America}
}

\author{Zheng Zeng}
\email{zhengzeng@ucsb.edu}
\orcid{0000-0001-9025-9427}
\affiliation{%
  \institution{University of California Santa Barbara}
  \city{Santa Barbara}
  \country{United States of America}
}
\affiliation{%
  \institution{NVIDIA}
  \city{Santa Clara}
  \country{United States of America}
}

\author{Fabrice Rousselle}
\email{frousselle@nvidia.com}
\orcid{0009-0003-2978-2130}
\affiliation{%
  \institution{NVIDIA}
  \city{Zurich}
  \country{Switzerland}
}

\author{Zahra Montazeri}
\email{zahra.montazeri@manchester.ac.uk}
\orcid{0000-0003-0398-3105}
\affiliation{%
  \institution{University of Manchester}
  \city{Manchester}
  \country{United Kingdom}
}

\author{Milo\v{s} Ha\v{s}an}
\email{milos.hasan@gmail.com}
\orcid{0000-0003-3808-6092}
\affiliation{%
  \institution{NVIDIA}
  \city{Santa Clara}
  \country{United States of America}
}

\renewcommand{\shortauthors}{Bowen Xue, Saeed Hadadan, Zheng Zeng, Fabrice Rousselle, Zahra Montazeri, and Milo\v{s} Ha\v{s}an}
\begin{abstract}
Creating photorealistic materials for 3D rendering requires exceptional artistic skill. Generative models for materials could help, but are currently limited by the lack of high-quality training data. While recent video generative models effortlessly produce realistic material appearances, this knowledge remains entangled with geometry and lighting. We present VideoNeuMat, a two-stage pipeline that extracts reusable neural material assets from video diffusion models. First, we finetune a large video model (Wan 2.1 14B) to generate material sample videos under controlled camera and lighting trajectories, effectively creating a "virtual gonioreflectometer" that preserves the model's material realism while learning a structured measurement pattern. Second, we reconstruct compact neural materials from these videos through a Large Reconstruction Model (LRM) finetuned from a smaller Wan 1.3B video backbone. From 17 generated video frames, our LRM performs single-pass inference to predict neural material parameters that generalize to novel viewing and lighting conditions. The resulting materials exhibit realism and diversity far exceeding the limited synthetic training data, demonstrating that material knowledge can be successfully transferred from internet-scale video models into standalone, reusable neural 3D assets. Code and data for this paper are at: \url{https://bowenxueai.github.io/VideoNeuMat/}. 
\end{abstract}




\begin{teaserfigure}
  \centering
  \includegraphics[width=\textwidth]{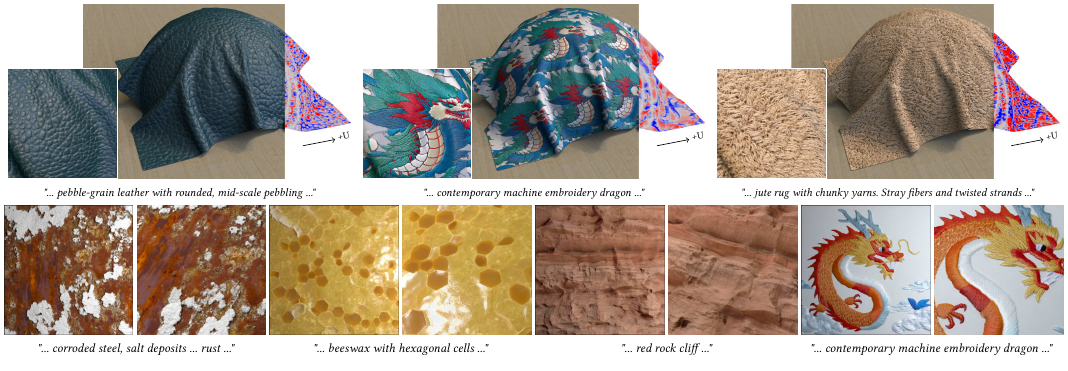}
  \caption{VideoNeuMat generates realistic and diverse neural materials, optionally tileable, from text or image prompts. The resulting material assets can be used on any surfaces under arbitrary lighting. We use the NeuMIP \cite{neumip} neural representation, which approximates parallax through neural offset (visualized as red and blue for negative and positive U-offset; V-offset looks similar). See supplementary materials for animated results.}
  \Description{Teaser description for accessibility.}
  \label{fig:teaser}
\end{teaserfigure}

\maketitle

\section{Introduction}
\label{sec:intro}

Creating realistic materials is critical for 3D rendering of believable virtual worlds, but requires time-consuming effort by highly skilled artists. There are few people in the world who can author truly convincing photo-real materials using classical tools such as procedural graphs, microfacet reflectance models, displacement mapping and layering; furthermore, the resulting complex layerings and graphs are often too costly for real-time rendering. Neural materials \cite{neumip,zeltner24} are a promising material deployment technology, relaxing some of the limitations of classical models, but they also lack straightforward solutions for photo-real content authoring.

On the other hand, recent image and video generative AI models effortlessly deliver virtually unlimited photo-real imagery and clearly understand material realism. To deliver a leap forward in the materials used in 3D rendering, we hypothesize that AI-assisted material generation can generate the quality of materials that currently only the best film material artists in the world can create. Ideally, this generative approach should directly produce neural representations, as the alternative of producing an open-ended number of layers using a large pallette of classical BRDF lobes appears harder, more constrained and less suitable for real-time applications.

However, the root problem is data. The straightforward approach of supervising the desired generative model directly by high quality materials hits a barrier: as noted by recent work on generating materials, both classical \cite{realmat} and neural \cite{gnm}, there are effectively zero datasets of materials at the realism level required. How do we overcome this data scarcity, and get to generative photo-real materials?

One approach would be to hire a team of the best material artists in the world, or build the largest and highest-quality material measurement lab. However, we believe there may a less brute-force solution. Video generative models already produce realistic material appearance, which suggests the possibility of "extracting" this material knowledge into standalone neural material assets, reusable on new shapes in new scenes. Of course, there is a technical obstacle: the materials in generated videos are entangled with shapes and lighting, and presented as final RGB frames.

In this paper, we show that neural materials can indeed be bootstrapped out of video models. Our solution is to teach a video model to generate videos of materials samples under a moving light and camera, and then train another model (also built on a video backbone) to turn such videos into neural materials.

First, we finetune a recent video model, Wan 2.1 14B \cite{wan2025}, to generate text- or image-conditioned videos of material samples under a specific light and camera trajectory. In a sense, we turn the video model into a "virtual gonioreflectometer". The finetuning is based on videos rendered using synthetic data from MatSynth \cite{vecchio2023matsynth}, with full displacement and multi-bounce path tracing. This teaches the video model the desired light/camera trajectory and the geometry of the sample, while preserving much of its material knowledge and realism. Therefore, we are able to generate videos of materials with appearances that go well beyond MatSynth in realism and diversity.

Second, we reconstruct a material in a neural representation previously used by \citet{neumip} and \citet{gnm}, consisting of a parallax module and main module, each defined by a feature texture, combined with a small neural network (MLP). The key challenge is that the videos generated in the first step are very small slices of the 6D reflectance space that neural materials model. While direct optimization of feature textures and MLPs to match the training views does give usable results in some cases, we also introduce a more robust and much faster reconstruction approach based on the idea of \emph{large reconstruction models} (LRMs) \cite{Hong2023LRMLR}. We finetune the transformer layers of a smaller video model, Wan 2.1 1.3B, to map 17-frame material videos into the feature textures of the corresponding material (with universal MLPs) in a single forward inference. The training uses a rendering loss under new view and lighting conditions, not known by the LRM, thus forcing it to produce neural materials that look plausible under a range of cameras and lights. In summary, our contributions are as follows:
\begin{itemize}
\item Teaching a large generative video model to become a "virtual gonioreflectometer", producing videos of material samples under a fixed light/camera trajectory, while preserving much of its prior knowledge about material realism.
\item Reconstructing neural materials from the above videos, usable on new shapes in new scenes, thus solving the challenge of disentangling the generative model's material knowledge from shape and lighting.
\item Introducing the first large reconstruction model (LRM) for producing neural materials (and materials in general) from sparse views, and also the first LRM finetuned from a video backbone, achieving significantly higher robustness and performance than direct optimization.
\end{itemize}

Our two-stage pipeline generates high-quality neural materials from text prompts, with realism and diversity far exceeding the small synthetic data used to train the method, thus breaking through the available data barrier (see \autoref{fig:teaser}). 

\begin{figure*}[t]
\centering
\includegraphics[width=\linewidth]{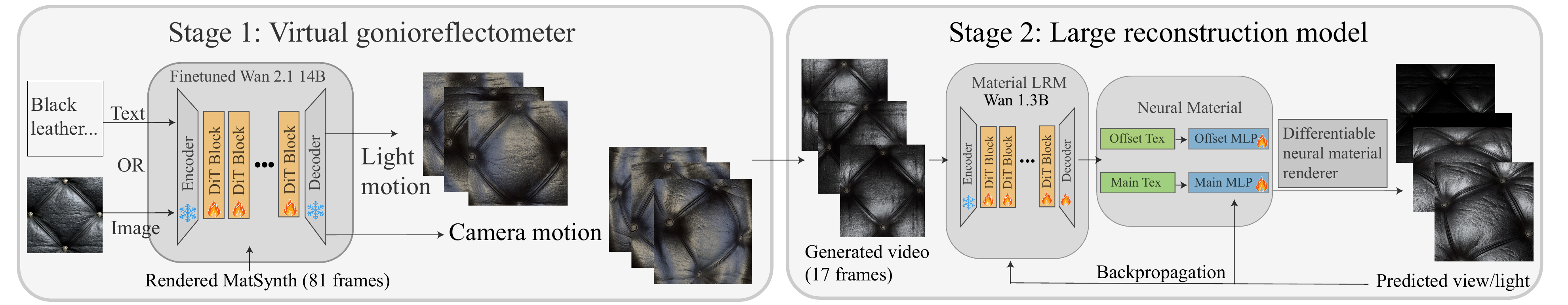}
\caption{Our method has two stages. First, we finetune a video diffusion model into a virtual gonioreflectometer that generates structured material videos from text or image prompts. Second, a feed-forward LRM infers a NeuMIP-style material from 17 frames using a rendering loss under novel views and lights. The resulting material supports relighting and novel shapes.}
\label{fig:pipeline}
\end{figure*}

\section{Related Work}
\label{sec:related}

\paragraph{Material representations.}
Realistic material appearance has been modeled through increasingly expressive representations, from analytic BRDF models~\cite{Reflectancemodelcook} to measured BTFs~\cite{10.1145/300776.300778} and spatially-varying BRDFs~\cite{10.1145/636886.636891}.
Neural representations have emerged as a compact alternative. \citet{https://doi.org/10.1111/cgf.13633} were the first to demonstrate that neural networks can faithfully fit and compress measured BTFs. NeuMIP~\cite{neumip} and follow-up work~\cite{zeltner24,Hierarchical} encodes view- and light-dependent appearance in learned latent textures decoded by small MLPs, achieving real-time performance independent of underlying material model complexity. We adopt the NeuMIP representation as our reconstruction target, as it balances expressiveness with efficient rendering, though any differentiable material representation would work with our LRM approach. 

\paragraph{Material capture.}
High-fidelity material capture traditionally requires specialized hardware such as gonioreflectometers~\cite{10.1145/3596711.3596750} or light stages~\cite{10.1145/3596711.3596762}. The concept of virtual gonioreflectometry dates back to at least
\citet{10.1145/37401.37434} and \citet{10.1145/142920.134075}, who simulated light interactions with a synthetic microstructure.
Single-image-based inverse rendering methods offer a more accessible alternative, learning neural priors to predict explicit SVBRDF maps~\cite{DADDB18,Shi2020,10.1145/3757377.3763848,10.1145/3610548.3618194}. A recent line of work relights the input image into new configurations to aid SVBRDF reconstruction~\cite{10.1145/3588432.3591515}, more recently leveraging diffusion priors~\cite{10.1145/3757377.3763809}. 
We take a different approach: instead of capturing real materials, we teach a video model to simulate the acquisition process, generating measurement-like sequences that can be converted to reusable neural assets, with no assumptions on analytic SVBRDF models. 

\paragraph{Material generation.}
Classical approaches synthesize materials through procedural graphs or example-based texture synthesis~\cite{10.1145/344779.345009}. More recently, GAN-based methods were introduced to generate directional and spatial material variation \cite{kuznetsov19,materialgan}.
Recent diffusion-based methods generate SVBRDF maps from text or images~\cite{ReflectanceFusion,vecchio2024matfuse,ControlMat}, but remain constrained by training data: existing material datasets such as MatSynth~\cite{vecchio2023matsynth} or Adobe Substance \cite{substance3d} lack the realism and diversity of natural imagery. PhotoMat \cite{zhou2023photomat} trains a material generator from a dataset of real images collected by the authors, but the size of such datasets is necessarily limited. The closest related work to ours \cite{gnm} generates neural materials in the NeuMIP representation using a diffusion model trained \emph{from scratch} and supervised directly by neural versions of materials from Adobe Substance; this data is not publicly available and is still limited. Our work sidesteps this data limitation by extracting material knowledge from video generative models pretrained on internet-scale data.

\paragraph{Video diffusion models.}
Large-scale video diffusion models have demonstrated remarkable ability to synthesize temporally coherent, photorealistic sequences \cite{blattmann2023stable,wan2025,kong2024hunyuanvideo}.
Recent work repurposes these models for 3D tasks: SV3D~\cite{sv3d} generates multi-view images for 3D reconstruction, while others extract motion priors for animation~\cite{hu2023animateanyone}.
VideoMat~\cite{https://doi.org/10.1111/cgf.70180} generates PBR materials for a given 3D mesh, leveraging a finetuned video diffusion model to produce a camera orbit, similar to our first stage but under uncontrolled natural lighting.

\paragraph{Large reconstruction models.}
Feed-forward reconstruction using LRMs~\cite{Hong2023LRMLR} has gained traction for 3D assets: many follow-up works predict neural radiance fields~\cite{li2023instant3d}, meshes \cite{xu2024instantmesh} or Gaussian splats~\cite{zhang2024gslrm} from sparse views in a single forward pass. To our knowledge, we introduce the first LRM for material reconstruction, mapping short material videos directly to neural material textures without iterative optimization; ours is also the first LRM that is trained from a strong video prior instead of from scratch.

\section{Structured Material Video Generation}
\label{sec:traj_video}

Extracting a neural material representation from a pretrained video generative model is a challenging task that we divide into two steps: structured video generation (this section) and neural material reconstruction (\autoref{sec:reconstruction}). 

Many material capture and generation pipelines start from a single photograph, under unknown or flash illumination. This is motivated by making material capture accessible to users without complex measurement setups.
As a result, view- and lighting-dependent effects (specular highlight motion, shadowing changes, and relief-induced parallax) are ambiguous: they are frequently entangled into a single 2D appearance map or suppressed by priors that favor view- and illumination-invariant outputs.
Since we are generating instead of capturing, we have the opportunity to produce multiple views and lighting conditions, simulating a more advanced capture setup.
We choose to represent a material sample as a short video, in which viewpoint and lighting changes become temporal variation.
This allows a pretrained video diffusion model to contribute its built-in temporal coherence and preserve its strong priors.


The rest of this section answers two questions: which video sequences are effective for making the subsequent reconstruction feasible, and how can we teach the model to generate such sequences without compromising its existing knowledge of realistic materials?

\subsection{Structured Video Trajectory}
\label{sec:trajectories}

\begin{figure}
\centering
\setlength{\tabcolsep}{2pt}
\begin{tabular}{@{}c@{\hspace{8pt}}c@{}}
\begin{subfigure}[b]{0.62\columnwidth}
\centering
\begin{tabular}{@{}cc@{}}
\small Phase 1: Fixed camera & \small Phase 2: Fixed light \\[2pt]
\includegraphics[
    trim={7.1cm 7.2cm 6.6cm 7.2cm},
    clip,
    width=0.48\linewidth
]{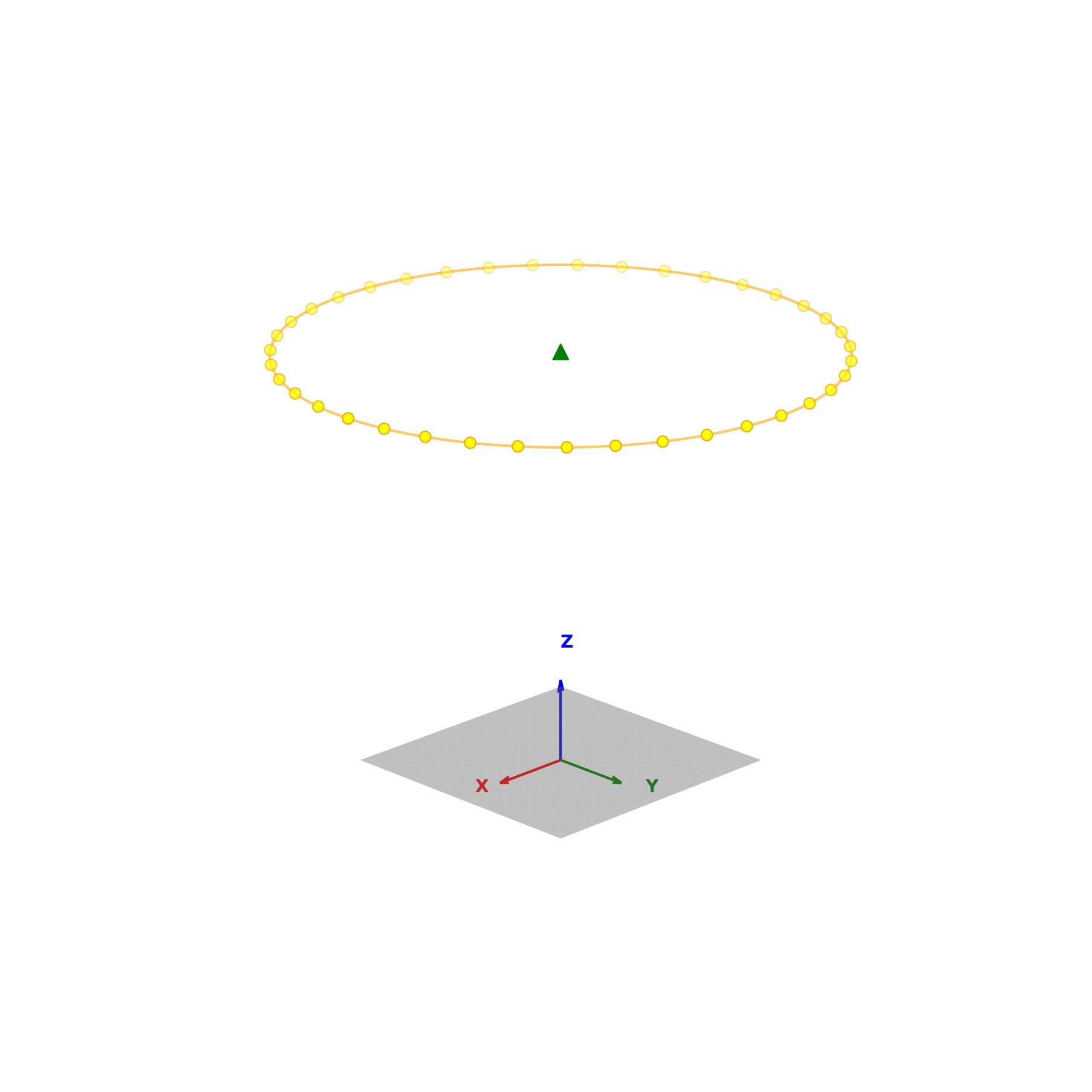} &
\includegraphics[
    trim={7.1cm 7.2cm 6.6cm 7.2cm},
    clip,
    width=0.48\linewidth
]{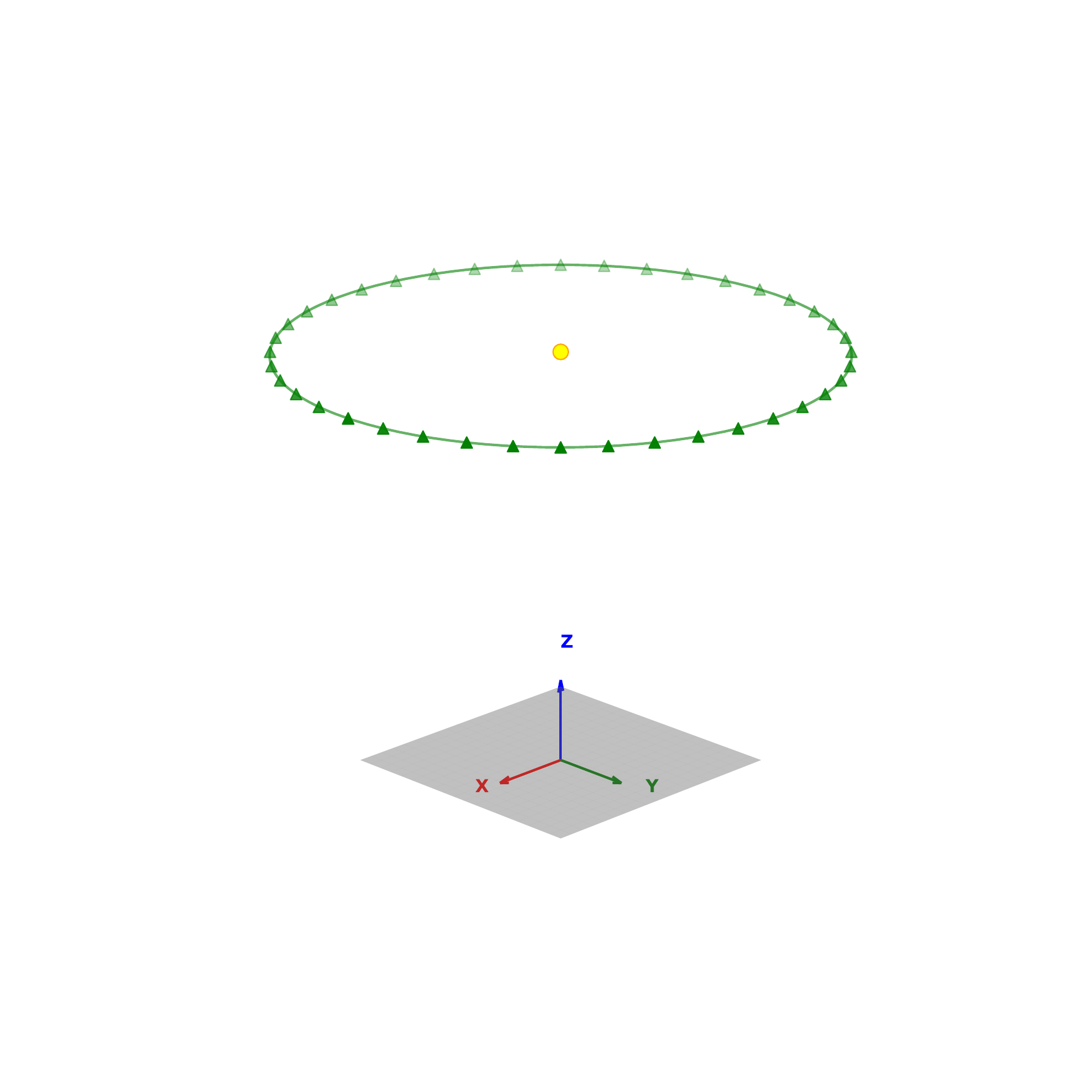}
\end{tabular}
\caption{Gonioreflectometer's trajectory}
\label{fig:gonio}
\end{subfigure}
&
\begin{subfigure}[b]{0.35\columnwidth}
\centering
\begin{tabular}{@{}c@{}}
\includegraphics[
    trim={8.1cm 12.3cm 7.3cm 7.2cm},
    clip,
    width=\linewidth
]{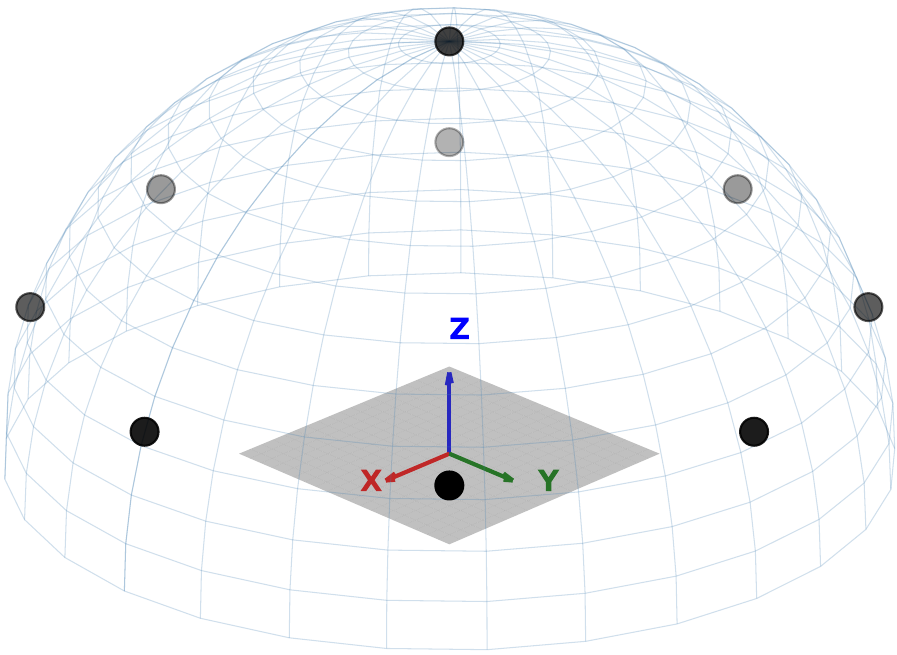}
\end{tabular}
\caption{LRM extra samples}
\end{subfigure}
\end{tabular}
\caption{Camera/light trajectories used for training. (a) Two-phase gonioreflectometer trajectory with 81 frames. (b) Additional 81 LRM training samples formed by cross-combining 9 camera and 9 light positions.}
\label{fig:trajectories}
\end{figure}

To make the generated path interpretable, we render training videos under controlled trajectories that separate the two sources of appearance variation: lighting and view.
Each sequence is composed of two consecutive, symmetric phases, as shown in \autoref{fig:gonio}: (1) Light-motion phase: the camera is fixed directly above the sample while the light moves along a circular path at an elevation angle of 56.31° ($= \arctan(3/2)$, corresponding to a height of $1.5$ units and radius of one unit above a unit-width sample); (2) Camera-motion phase: the light is fixed directly above while the camera orbits along a circular path at the same elevation angle. 

%
By separating the two phases, we provide clean supervision that is easy for the video model to learn: the 
light-motion phase teaches illumination response, while the camera-motion 
phase teaches view-dependent effects and parallax.

\def\reconNumcols{10}

\ifdefined\reconImgwidth\else\newlength{\reconImgwidth}\fi
\setlength{\reconImgwidth}{\dimexpr(\textwidth / \reconNumcols) - 2pt\relax}

\def\reconBasepath{image/reconstruction/diffusion}

\def\reconClothframe{0008}
\def\reconClothframeB{0007}
\def\reconPlanetrain{0}

\def\reconImg#1{%
    \IfFileExists{#1}{%
        \includegraphics[width=\reconImgwidth]{#1}%
    }{%
        \includegraphics[width=\reconImgwidth]{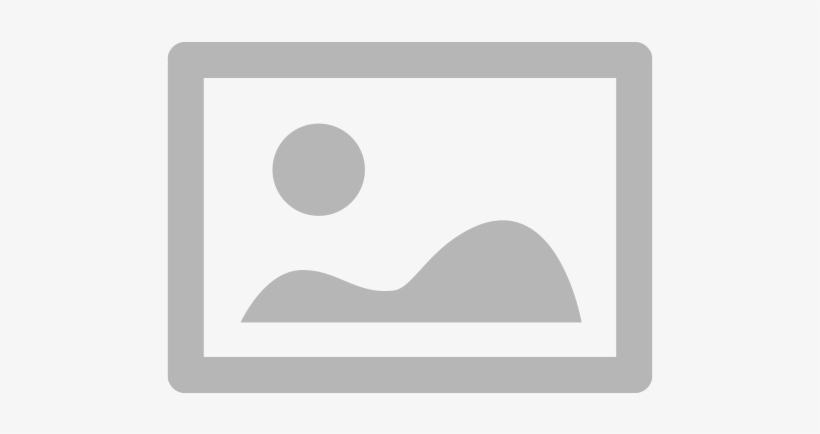}%
    }%
}

\def\reconClothImg#1#2{\reconBasepath/cloth/#1/image_#2.jpg}
\def\reconClothImgB#1#2{\reconBasepath/cloth/#1/image_#2.jpg}

\def\reconGTImg#1#2{\reconBasepath/plane/#1/gt_train_#2.jpg}
\def\reconPlaneTrainImg#1#2{\reconBasepath/plane/#1/train_views_#2.jpg}
\def\reconPlaneTrainShiftX#1#2{\reconBasepath/plane/#1/train_offset_shift_x_#2.jpg}

\def\reconColDiffusion{Diffusion}
\def\reconColReconstruction{Reconstruction}
\def\reconColOffset{Offset}
\def\reconColCurved{Curved surf.}

\def\reconMatA{basket}
\def\reconPromptA{Woven basket texture}

\def\reconMatB{hives}
\def\reconPromptB{Honeycomb hive pattern}

\def\reconMatC{0004_seed42_full_rectify11000complex1k}
\def\reconPromptC{"... Button-tufted leather upholstery ..."}

\def\reconMatD{tree}
\def\reconPromptD{Tree bark texture}


\def\reconMatBlock#1{%
    \reconImg{\reconGTImg{#1}{\reconPlanetrain}} &
    \reconImg{\reconPlaneTrainImg{#1}{\reconPlanetrain}} &
    \reconImg{\reconPlaneTrainShiftX{#1}{\reconPlanetrain}} &
    \reconImg{\reconClothImg{#1}{\reconClothframe}}&
    \reconImg{\reconClothImgB{#1}{\reconClothframeB}}%
}

\def\reconMakerow#1#2#3#4{%
    \reconMatBlock{#1} & \reconMatBlock{#3} \\
}

\begin{figure*}[t]
    \centering
    \setlength{\tabcolsep}{1pt}
    \begin{tabular}{ccccc|ccccc}
        \small\reconColDiffusion & 
        \small\reconColReconstruction & 
        \small\reconColOffset & 
        \small\reconColCurved &
        \small\reconColCurved &        
        \small\reconColDiffusion & 
        \small\reconColReconstruction & 
        \small\reconColOffset & 
        \small\reconColCurved &        
        \small\reconColCurved \\[2pt]
        \hline \\[-6pt]
        \reconMakerow{\reconMatA}{\reconPromptA}{\reconMatB}{\reconPromptB}
        \reconMakerow{\reconMatC}{\reconPromptC}{\reconMatD}{\reconPromptD}%
    \end{tabular}
    \caption{For a few example materials, we show the first generated frame, the LRM-reconstructed rendering of the same frame, the U-offset map, and renderings of the material on a curved surface under different environment illuminations. The offset amount (for the neural parallax mapping effect of NeuMIP) is shown along the U-axis in texture space. A red pixel means texture look-up a few pixels to the left, while blue means offset to the right). Not only our approach can reconstruct the generated frames faithfully, but also it generates meaningful offset maps, indicating a correct understanding of the material geometry.}
    \label{fig:recon_results}
\end{figure*}

\paragraph{Perspective projection and near-field lighting.}
Traditional BTF acquisition uses near-orthographic cameras and typically rectifies 
data such that each pixel consistently maps to the same surface point 
across all view / light conditions. This layout simplifies data processing, but such imagery does not exist in natural videos and falls far outside the pretraining 
distribution of video diffusion models.
We therefore adopt standard perspective projection and a small near-field area light, which produces natural patterns that better match real video statistics.
Although this sacrifices per-pixel alignment in the view variation phase, we show in 
\autoref{sec:ablation_trajectory} that the preserved prior 
leads to higher-quality generated materials and better downstream 
reconstruction.



\subsection{From Pretrained Video Model to Material Model}
\label{sec:finetune}


Our intent is not to rebuild video realism from scratch, but to steer a strong pretrained model towards producing material sample videos under our controlled measurement trajectories. Our training data (MatSynth renderings) is limited in realism and diversity compared to what the base model has implicitly learned; successful finetuning should carefully retain the pretrained material realism while only teaching the base model the new trajectory and sample geometry.

\paragraph{Synthetic data generation.} 
We render training videos from MatSynth using Mitsuba 3, utilizing over 5,600 materials at 1024$\times$1024 resolution. 
To capture realistic parallax and self-shadowing, we apply the 
height maps as true geometric displacement on a 1024$\times$1024 mesh.
We augment the dataset with 90° rotations of each material. We apply a distance-based light intensity correction to the synthetic training data so that overall brightness remains constant.

\paragraph{Caption generation.}
Since MatSynth does not provide text captions, we generate them from 
rendered training videos using Qwen2-VL~\cite{Qwen2VL}.
Since the captioner naturally describes visible lighting variations, we 
explicitly prevent lighting-related text content through two strategies:
\begin{enumerate}
\item Filtered captions: we prompt the model to describe the 
video freely, then post-process to remove sentences containing 
standalone lighting-related words (e.g., "light", "lighting", 
"illumination");
\item Material-only captions: we directly instruct the captioner 
to focus only on intrinsic material attributes and ignore all lighting information.
\end{enumerate}
During training, we mix both caption types equally.

\paragraph{Full finetuning and LoRA}  We choose a recent open-source model, Wan2.1-14B, due to its strong generative prior and clean software implementation in DiffSynth-Studio~\cite{diffsynthstudio}. We finetune the model using the standard diffusion objective of the base model, without introducing task-specific losses. We explored two finetuning strategies with different prior-preservation characteristics: LoRA and full finetuning, both of which lead to realistic results.  LoRA (rank 32) updates only low-rank adapters, which in theory better preserve the pretrained prior. However, this comes at a trade-off: LoRA achieves weaker trajectory compliance (see Table~\ref{evaluation}b), suggesting the measurement trajectory requires broader weight modifications. For full finetuning, we apply classifier-free guidance dropout (probability 0.1) to preserve unconditional generation capability, ensuring negative prompts remain effective; LoRA inherently retains this capability without explicit dropout. We use both model types in our results.



\paragraph{Image conditioning.} We also finetune an image-conditioned variant that generates from a reference photograph.
For training, we render MatSynth materials under natural illumination sampled from 500 environment maps (normalized to equivalent irradiance), encouraging disentanglement of intrinsic material appearance from incident lighting (\autoref{fig:i2v_results}).


\definecolor{bestgreen}{RGB}{34, 139, 34}      
\definecolor{secondgreen}{RGB}{144, 238, 144}  
\definecolor{worstred}{RGB}{220, 53, 69}       
\definecolor{secondorange}{RGB}{255, 165, 0}   

\newcommand{\best}[1]{\cellcolor{bestgreen!40}\textbf{#1}}
\newcommand{\second}[1]{\cellcolor{secondgreen!40}#1}
\newcommand{\worst}[1]{\cellcolor{worstred!40}\textbf{#1}}
\newcommand{\secondworst}[1]{\cellcolor{secondorange!40}#1}

\begin{table}[t]
\centering
\caption{Ablation study results. \colorbox{bestgreen!40}{\textbf{Best}}, \colorbox{secondgreen!40}{2nd best}, \colorbox{secondorange!40}{2nd worst}, \colorbox{worstred!40}{\textbf{worst}}. "Rand. init." is short for random initialization.}
\label{tab:ablation}
\footnotesize
\setlength{\tabcolsep}{2pt}

\begin{tabular}{l|cccc|cccc}
\toprule
& \multicolumn{4}{c|}{\textbf{Features}} & \multicolumn{4}{c}{\textbf{Metrics}} \\
\cmidrule(lr){2-5} \cmidrule(lr){6-9}
\textbf{Experiment} & 
\rotatebox{70}{\scriptsize Pretrained WAN} & 
\rotatebox{70}{\scriptsize Pretrained MLP} & 
\rotatebox{70}{\scriptsize Trainable MLP} & 
\rotatebox{70}{\scriptsize VAE upsampler} & 
LPIPS $\downarrow$ & 
MAPE $\downarrow$ & 
MSE $\downarrow$ & 
PSNR $\uparrow$ \\
\midrule

Full method & 
\checkmark & \checkmark & \checkmark & \checkmark &
0.3017 & \best{0.0407} & \best{0.005667} & \best{24.29} \\

Linear upsampler & 
\checkmark & \checkmark & \checkmark & &
\best{0.2919} & 0.0454 & \second{0.005767} & \second{24.25} \\

Frozen MLP & 
\checkmark & \checkmark & & \checkmark &
0.3155 & 0.0461 & 0.006105 & 23.96 \\

Rand.\ init.\ MLP & 
\checkmark & & \checkmark & \checkmark &
0.3341 & \worst{0.0565} & \worst{0.009961} & \worst{22.13} \\

Rand.\ init.\ LRM & 
& \checkmark & \checkmark & \checkmark &
\worst{0.3543} & \secondworst{0.0563} & \secondworst{0.008769} & \secondworst{22.45} \\

\midrule
\multicolumn{9}{c}{\textit{Network size (base: 64$\times$4)}} \\
\midrule

128$\times$6 & 
\checkmark & & \checkmark & \checkmark &
0.3211 & 0.0483 & 0.007285 & 23.22 \\

32$\times$2 & 
\checkmark & & \checkmark & \checkmark &
0.3325 & 0.0501 & 0.007287 & 23.17 \\

\midrule
\multicolumn{9}{c}{\textit{LRM input views (base: 17 frames)}} \\
\midrule

5 inputs & 
\checkmark & \checkmark & \checkmark & \checkmark &
\secondworst{0.3352} & 0.0517 & 0.006763 & 23.45 \\

41 inputs & 
\checkmark & \checkmark & \checkmark & \checkmark &
\second{0.3009} & \second{0.0435} & 0.006151 & 24.04 \\

\bottomrule
\end{tabular}

\scriptsize
\textit{Note: Checkmarks indicate features present. Full method uses 64$\times$4 MLP and LRM 17 input frames.}

\end{table}

\begin{figure}[tb]
  \centering
  \begin{subfigure}[t]{0.495\columnwidth}
    \centering
    \includegraphics[trim={0pt 0pt 0pt 700pt}, clip, width=\linewidth]{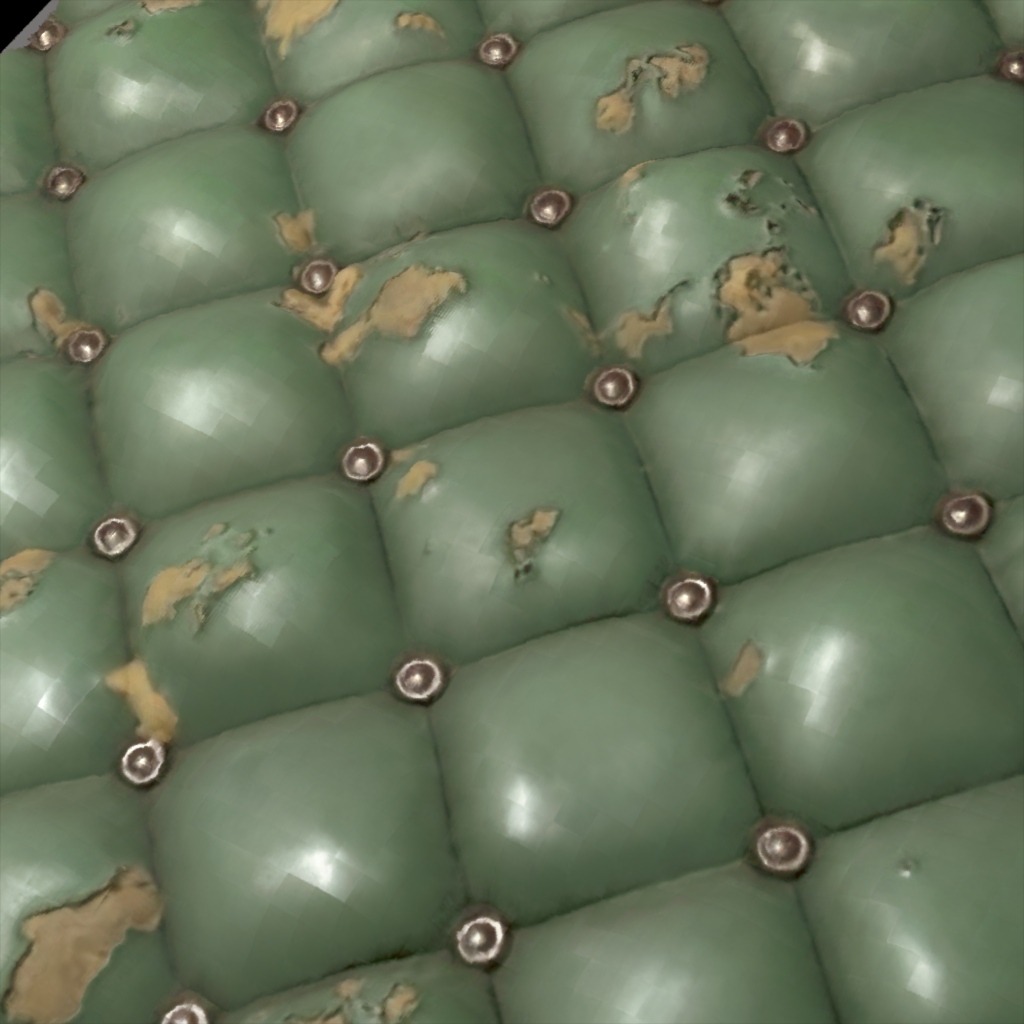}
  \end{subfigure}
  \begin{subfigure}[t]{0.495\columnwidth}
    \centering
    \includegraphics[trim={0pt 0pt 0pt 700pt}, clip, width=\linewidth]{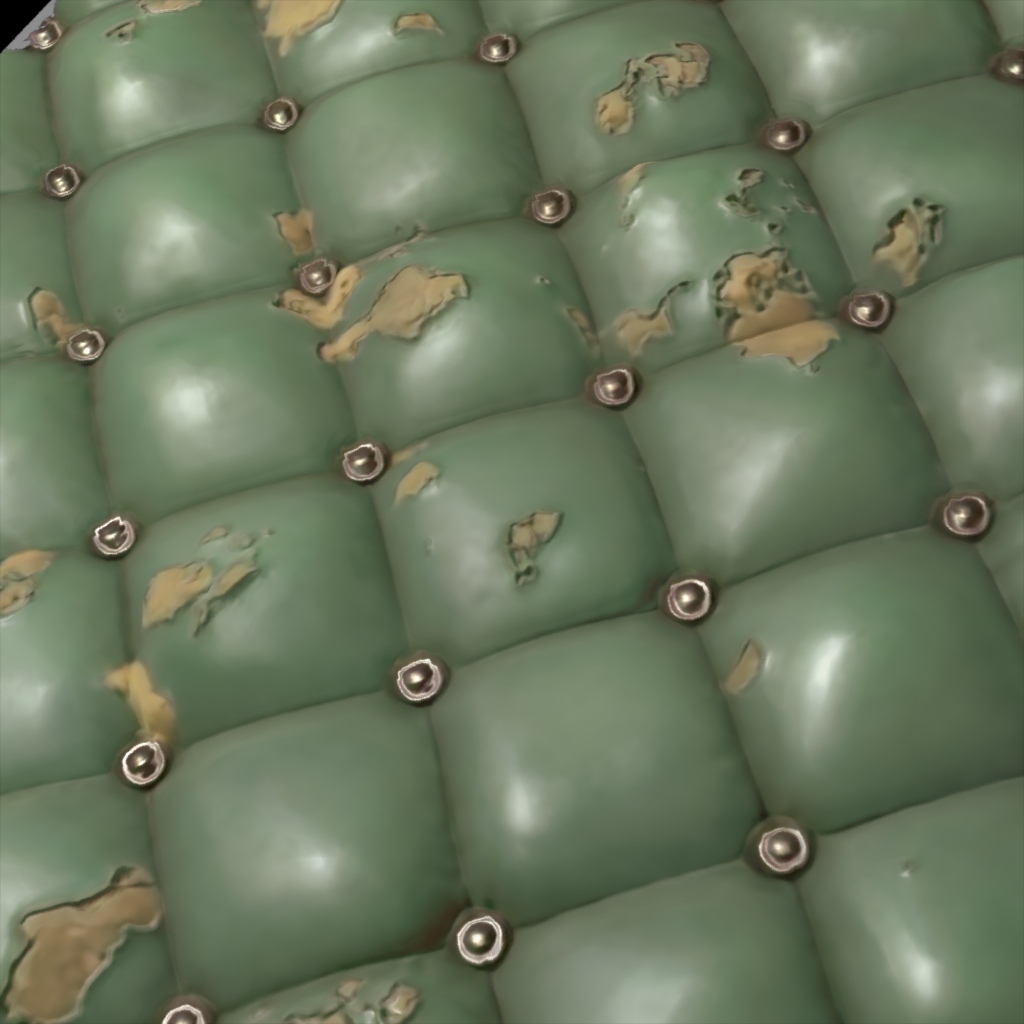}
  \end{subfigure}
  \begin{subfigure}[t]{0.495\columnwidth}
    \centering
    \includegraphics[trim={0pt 0pt 0pt 700pt}, clip, width=\linewidth]{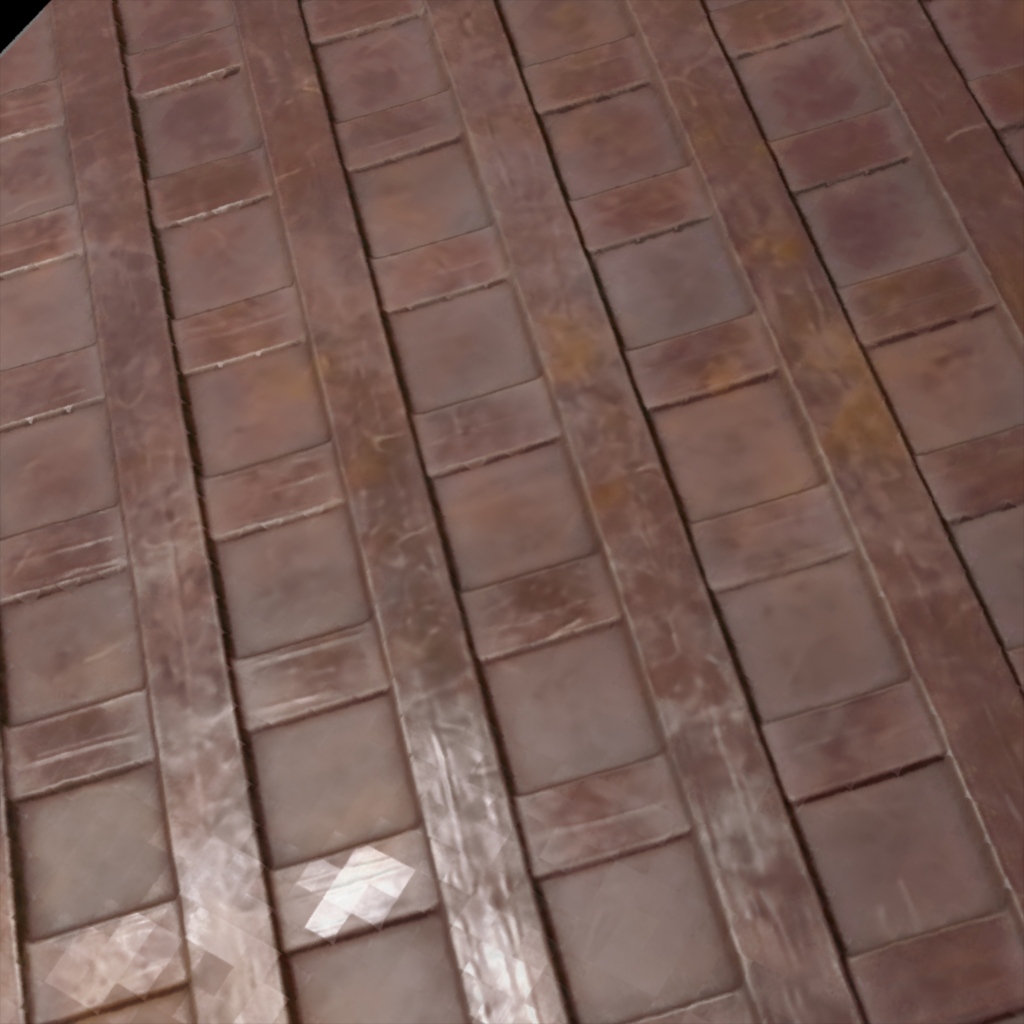}
    \caption{Linear layer}
  \end{subfigure}
  \begin{subfigure}[t]{0.495\columnwidth}
    \centering
    \includegraphics[trim={0pt 0pt 0pt 700pt}, clip, width=\linewidth]{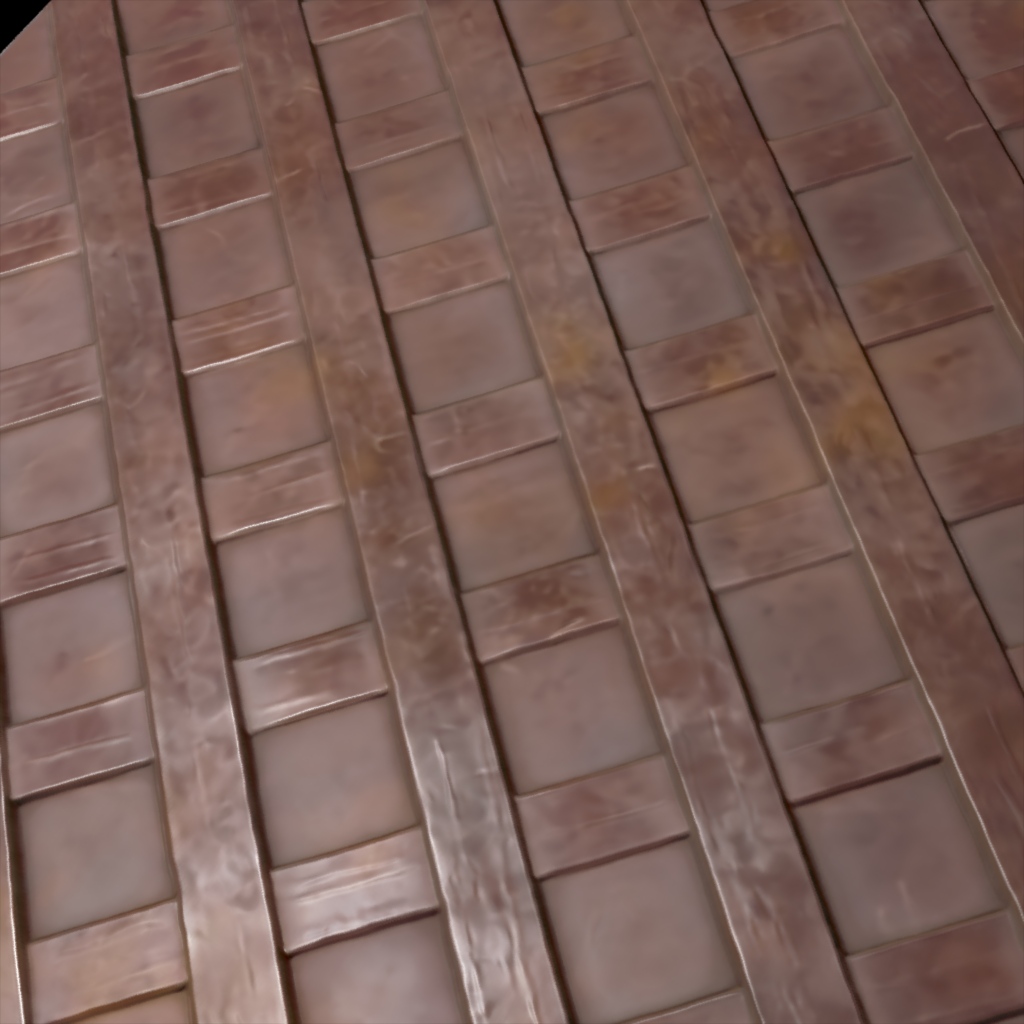}
    \caption{VAE decoder}
  \end{subfigure}


  \caption{\textbf{LRM upsampler}. A linear layer as the upsampler of the DiT tokens processes each token separately from the others, causing patch artifacts, while the convolutional layers in the VAE decoder resolve this issue.}
  \label{fig:upsampling}
\end{figure}

\paragraph{Training details.} We finetune Wan2.1-14B updating only the diffusion transformer (DiT) while keeping the variational autoencoder and text encoder (UMT5-XXL) frozen.
We train models at $512 \times 512$ with 81-frame sequences (the native video length of the Wan model), using AdamW with a learning rate of $1 \times 10^{-5}$.
Training is conducted on 4 NVIDIA A100 GPUs.
For the image-conditioned variant, we finetune Wan2.1-I2V-14B-480P with identical hyperparameters, conditioning on a reference material photograph.

\paragraph{Tileable material variant.}
For applications requiring tileable materials, we further finetune our model on the tileable subset.
This restricts the training set to roughly 3,000 materials with periodic boundary conditions. The resulting model generates outputs that are largely tileable, though with occasional imperfect samples. We also apply a latent-space post-processing step, which modifies a narrow band along the boundaries to reduce seam visibility. See the supplementary document for details.

\subsection{Inference and Evaluation}

\paragraph{Inference with novel prompts.}
Although training captions originate from synthetic MatSynth renderings, at inference the finetuned model can be driven by prompts that are outside the training caption distribution, producing material appearances that go beyond MatSynth in both diversity and realism.

\paragraph{Prompt expansion.}
We use an LLM (Claude Opus 4.5 \cite{claude2024}) to expand short material prompts into detailed captions matching the training format, adding surface micro-geometry, weathering, color variation, and other material-specific cues. Examples are provided in the supplementary material.

\paragraph{Flexible inference resolution.}
Thanks to the DiT architecture of Wan2.1, our finetuned model exhibits resolution flexibility beyond the training resolution.
We train model at $512 \times 512$, but inference at resolutions up to $1920 \times 1920$ without architectural modification, enabling high-resolution material videos with fine detail suitable for production use cases.

\paragraph{BTF sequence quality metric.}
Since no existing metric quantifies BTF sequence quality, we propose \emph{RPC (Residual Parallax Coherence)}.
The key insight is that for a perfectly rectified BTF sequence, residual optical flow should correlate monotonically with camera baseline: larger pose changes induce proportionally larger parallax for non-planar surface details.

We first obtain rectified frames by unwarping rendered images to a canonical UV space using reference-plane position AOV to remove perspective distortion.
On Phase~2 (the camera-motion phase), we form frame pairs $(t, t+\Delta)$ with multiple temporal lags $\Delta$ to cover a wide range of pose baselines.
For each pair, we compute a composite baseline combining rotation and normalized translation:
\begin{equation}
    b_{t,\Delta} = \sqrt{\Delta\theta_{t,\Delta}^2 + \left(\frac{\|\Delta\mathbf{t}_{t,\Delta}\|}{s}\right)^{\!2}},
\end{equation}
where $\Delta\theta_{t,\Delta}$ is the relative rotation angle, $\Delta\mathbf{t}_{t,\Delta}$ is the translation difference, and $s$ is the world-space extent of the material sample.
We then measure the median optical-flow magnitude $m_{t,\Delta}$ over a robust intensity mask that excludes the 10th/90th intensity percentiles after bilateral filtering.
RPC loss penalizes weak monotonic dependence:
\begin{equation}
    L_{\text{rpc}} = 1 - \max\!\bigl(0,\,\rho_s(\{b_{t,\Delta}\},\{m_{t,\Delta}\})\bigr),
\end{equation}
where $\rho_s$ is Spearman correlation, chosen over Pearson for robustness to outliers and to capture monotonic relationships.
Optical flow is computed using Dual TV-L1~\cite{10.1007/978-3-540-74936-3_22}.

\section{Neural Material Reconstruction}
\label{sec:reconstruction}
In this section, we discuss ways to obtain a neural representation from the videos of our gonioreflectometer. Specifically, the goal is to obtain the parameters of a given neural material model (in this case NeuMIP) from a set of frames under known geometry, camera pose, and illumination. Namely, we will discuss \emph{direct optimization}, and a data-driven approach inspired by \emph{Large Reconstruction Models (LRMs)} \cite{Hong2023LRMLR}. Despite our focus on NeuMIP, the approach is applicable to any neural (or even classical) material model with a differentiable rendering operator available.

\subsection{NeuMIP representation}

Instead of the original NeuMIP formulation \cite{neumip}, we use the recent modification by \citet{gnm}, which does not use a multi-resolution pyramid, and uses universal instead of per-material MLPs. Our representation can be easily converted into the original one by distillation if desired. 

This architecture has four components: a per-material offset feature texture $\toff$, a per-material reflectance feature texture $\trgb$, a universal offset MLP $\moff$, and a universal reflectance MLP $\mrgb$. These can be thought of as an offset module (responsible for the parallax effect) and a reflectance module (returning the final RGB reflectance value). 

Given a material-space location $\bu$, incoming (light) direction $\bi$ and outgoing (camera) direction $\bo$, the final reflectance value $f_{\phi,\theta}(\bu, \bi, \bo)$, where $\phi$ are the MLP parameters and $\theta$ are the feature texture parameters, is evaluated as follows:
\begin{alignat}{3}
    \Delta \bu &= \moff(\toff[\bu], \bo) &&\in [-1,1]^2,  \\
   f_{\phi,\theta}(\bu, \bi, \bo) &= \mrgb(\trgb[\bu + \Delta \bu], \bi, \bo) &&\in \mathbf{RGB}. 
\end{alignat}
This evaluates the offset module at $\bu$, computes an offset $\Delta \bu$, and evaluates the reflectance module at $\bu + \Delta \bu$. Feature texture queries are by bilinear interpolation. See \citet{gnm} (Section 3) for more details. We require a rendering operator (differentiable in $\phi,\theta$) that evaluates the model on a flat plane with a perspective camera and a point light, which is straightforward to implement.

\subsection{Direct optimization}
\label{sec:direct-fit}
A straightforward approach to obtain a neural representation from a sequence of frames under known geometry and lighting is to directly optimize the parameters of the neural material in an inverse rendering framework. This involves differentiably rendering candidate images of the neural material and comparing them to the input frames using an image-based loss. The parameters of the neural model are then iteratively updated using gradient descent. Our experiments show direct optimization can sometimes produce good results, but with three significant limitations.



\paragraph{Limitations.} Direct optimization requires perfect multi-view consistency; otherwise, conflicting information prevents convergence to a sharp solution. Therefore, it works best on traditionally rendered videos. Since the videos generated by the diffusion model have a degree of view/light inconsistency, the reconstruction has significant artifacts as shown in \autoref{fig:direct_optim}. Worse, direct optimization motivates the neural model to only explain the input video sequence. Since the video covers a limited slice of the 4D space $(\bi, \bo)$, and the optimization does not make use of any prior knowledge, the resulting neural material can fail to robustly generalize to unseen camera/light locations (\autoref{fig:direct_optim}). Lastly, direct optimization is slow due to iterative rendering and optimization per-material (1.5 hours to fit 81 frame videos at a resolution of 1024). 


\subsection{Large Reconstruction Model}
\label{sec:lrm}
To overcome the limitations of direct optimization, we propose a feed-forward Large Reconstruction Model (LRM) that replaces per-material, iterative optimization with a single forward pass: given a material video in the gonioreflectometer's camera/light trajectory as input, the LRM outputs the latent textures of our neural representation. Our LRM uses only 17 input frames, a subset of the 81 generated by the video model. Despite the sparse inputs, the LRM is much more robust compared to direct optimization, both for synthetic and especially for diffusion-generated materials (\autoref{fig:direct_optim}). It is also much faster than direct optimization (roughly 4 seconds per material at resolution 1024). \autoref{fig:recon_results} shows the LRM approach can successfully reconstruct the material appearance and geometry (through the offset module).

\paragraph{Training.} We do not directly supervise the LRM output using a set of ground truth latent textures, as we do not have access to such a dataset (\citet{gnm} did not release their data). Instead, we use a rendering loss: we render images using the latent textures for new view / lighting conditions unknown to the LRM, and compute an L2 loss against the ground truth MatSynth rendering. This rendering loss is calculated not only on the gonioreflectometer's trajectory (81 frames), but also on new combinations of light and camera locations (81 frames), \autoref{fig:trajectories}. The latter set consists of camera and light locations placed on the hemisphere on center texel, ensuring 3 evenly spaced samples in each dimension of the 4D space of $w_i$, $w_o$, thus $3^4=81$ frames.

\paragraph{LRM architecture.}
We adopt the Wan2.1-1.3B video diffusion model and repurpose it into a non-generative regressor that converts a material video into NeuMIP latent textures. The DiT operates in Wan's VAE latent space, which we reuse. We encode the input frames using the frozen VAE encoder before feeding them to the DiT. The DiT output tokens corresponding to the first frame of the video are then fed to the VAE decoder (with modified final layers) to obtain the NeuMIP latent textures $\toff$ and $\trgb$ (12 channels each). All other output tokens are ignored (i.e., do not affect loss computation). The DiT and VAE decoder are trained end-to-end.

We initialize both the DiT and VAE decoder weights using the pretrained Wan weights as opposed to random initialization. As shown in Table \ref{tab:ablation}, this produces better results. We attribute this to the pretrained model's ability to parse temporal appearance variation from input videos, a skill that transfers directly to material reconstruction. The only layers that are randomly initialized are the DiT and VAE heads (final layers), due to the domain gap.

\paragraph{NeuMIP MLP decoders}
The textures $\toff$ and $\trgb$ produced by the LRM are fed into the offset decoder $\moff$ and reflectance decoder $\mrgb$ to render images under novel view / lighting. The MLPs have 4 hidden layers with 64 neurons and are trained end-to-end with the LRM. The MLPs can be randomly initialized, or pretrained on a set of materials. The latter produces better results, as shown in \autoref{tab:ablation}.



\paragraph{Input frame count.}
Video diffusion VAEs typically apply a temporal downsampling factor of 4 after an 
independent keyframe, requiring frame counts of the form $1+4n$, 
hence our choices of 5, 17, 41, and 81 frames. Our two-phase trajectory naturally suggests a minimum sampling density: 
enough frames to capture both illumination response (light-motion phase) 
and view-dependent effects (camera-motion phase).
We find 17 uniformly-spaced frames optimal; reducing to 5 frames 
undersamples the light-motion phase and degrades perceptual quality (LPIPS), 
while 41 frames offers no improvement and slightly hurts generalization, 
likely due to overfitting to redundant observations.

\paragraph{Training details.} We pretrain universal NeuMIP MLPs on 512 random MatSynth materials, then jointly train with the LRM on MatSynth renderings. We train the LRM on MatSynth renderings at resolution $256^2$ for 75 hours at fp32 precision, followed by 60 hours at bf16 at resolution $1024^2$, using 64 A100 GPUs. In addition to L2 pixel-wise loss, we used LPIPS with weight 0.1. 

\section{Results}
\label{sec:results}

\autoref{fig:recon_results_2} shows the final output of our pipeline, i.e., generated neural materials from various prompts, rendered on flat and curved geometries. See supplementary for more results.

\begin{figure*}[!bth]
\centering
\includegraphics[width=\textwidth]{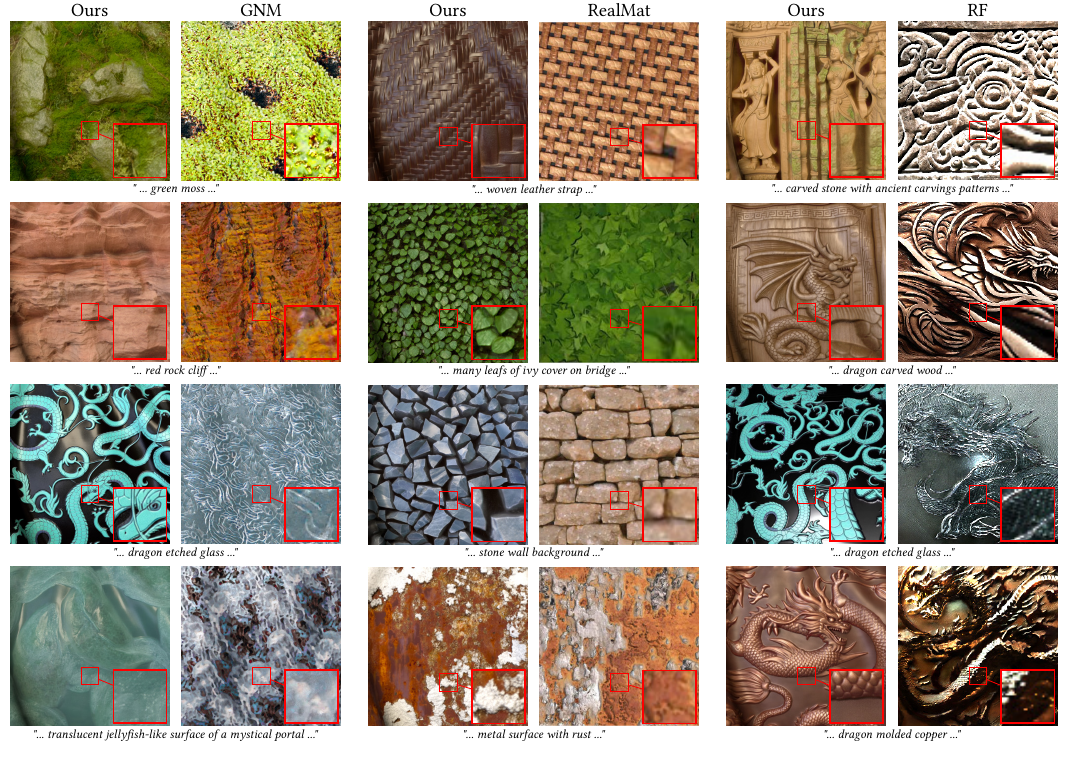}
\caption{\textbf{Qualitative comparison with baseline methods:} GNM \cite{gnm}, RealMat \cite{realmat} and ReflectanceFusion \cite{ReflectanceFusion}. All three previous methods are based on strong diffusion backbones, and produce meaningful results. In our opinion, our results are the most successful at extracting realistic material appearance from the backbone generative model. \newline\footnotesize{Credits: GNM panels \textcopyright{} Generative Neural Materials project authors; RealMat panels \textcopyright{} 2026 Eurographics - The European Association for Computer Graphics and John Wiley \& Sons Ltd.; ReflectanceFusion panels \textcopyright{} 2024 The Authors.}}
\label{fig:compare_all}
\end{figure*}

\paragraph{Qualitative comparison.} We compare against three baselines: GNM \cite{gnm}, RealMat~\cite{realmat}, and ReflectanceFusion~\cite{ReflectanceFusion} in \autoref{fig:compare_all}. We use examples from their original publications and generate results with the same text prompts. Note that viewing and lighting conditions differ across methods. Our approach consistently achieves higher output resolution and more pronounced displacement effects in all comparisons.

\paragraph{Single-image-to-material comparison.}
  We further evaluate image-conditioned material generation on 7 materials from Adobe Substance \cite{substance3d}, not used in training.
  Given a single input image, we compare against CHORD \cite{10.1145/3757377.3763848} and MatFusion \cite{10.1145/3610548.3618194}.
  As shown in Figure~\ref{fig:i2t_comparison}, our method better preserves the material
  structure and relief cues across diverse materials. We report quantitative results in  \autoref{tab:i2t_comparison}, evaluating both the displayed frame and additional test views from the same materials. High-relief materials are particularly challenging for competing SVBRDF methods, especially since the predicted normal maps were not integrated into a height field to capture parallax effects. Note that our approach is a generative model conditioned on the input image, so some differences in color and tone are expected; still, our method achieves the lowest error across all metrics.
%

\providecommand{\iTwoTCompareBase}{i2vcompare/}
\providecommand{\iTwoTPlaceholderPath}{placeholder.jpg}

\newcommand{\iTwoTMCellHeight}{0.136\textwidth}
\newcommand{\iTwoTMInputWidth}{0.13\textwidth}
\newcommand{\iTwoTMResultWidth}{0.174\textwidth}
\newcommand{\iTwoTMHeaderHeight}{0.020\textwidth}

\newcommand{\iTwoTMFixedBox}[2]{%
  \parbox[c][\iTwoTMCellHeight][c]{#1}{\centering #2}%
}
\newcommand{\iTwoTMHeader}[2]{%
  \parbox[c][\iTwoTMHeaderHeight][c]{#1}{\centering\scriptsize\textbf{#2}}%
}
\newcommand{\iTwoTMInput}[1]{%
  \iTwoTMFixedBox{\iTwoTMInputWidth}{%
    \includegraphics[height=0.13\textwidth,angle=-90,origin=c]{\iTwoTCompareBase#1}%
  }%
}
\newcommand{\iTwoTMResult}[1]{%
  \iTwoTMFixedBox{\iTwoTMResultWidth}{%
    \includegraphics[height=0.13\textwidth,trim=0 0 0 256bp,clip]{\iTwoTCompareBase#1}%
  }%
}

\begin{table*}[htbp]
\centering
\begin{minipage}[t]{0.72\textwidth}
\centering
\setlength{\tabcolsep}{0pt}
\renewcommand{\arraystretch}{1.0}
\begin{tabular}{@{}c@{\hspace{1pt}}c@{\hspace{1pt}}c@{\hspace{1pt}}c@{\hspace{1pt}}c@{\hspace{1pt}}c@{}}
\iTwoTMHeader{\iTwoTMInputWidth}{Input} &
\iTwoTMHeader{\iTwoTMResultWidth}{CHORD} &
\iTwoTMHeader{\iTwoTMResultWidth}{MatFusion} &
\iTwoTMHeader{\iTwoTMResultWidth}{Our full} &
\iTwoTMHeader{\iTwoTMResultWidth}{GT} &
\iTwoTMHeader{\iTwoTMResultWidth}{Our LRM} \\

\iTwoTMInput{scifi_weaved_carbon_fiber/reference_firstframe.jpg} &
\iTwoTMResult{scifi_weaved_carbon_fiber/frame060/scifi_weaved_carbon_fiber_chord_frame060_src059.jpg} &
\iTwoTMResult{scifi_weaved_carbon_fiber/frame060/scifi_weaved_carbon_fiber_matfusion_frame060_src059.jpg} &
\iTwoTMResult{scifi_weaved_carbon_fiber/frame060/scifi_weaved_carbon_fiber_our_full_frame060_src236.jpg} &
\iTwoTMResult{scifi_weaved_carbon_fiber/frame060/scifi_weaved_carbon_fiber_gt_frame060_src059.jpg} &
\iTwoTMResult{scifi_weaved_carbon_fiber/frame060/scifi_weaved_carbon_fiber_our_lrm_frame060_src236.jpg} \\

\iTwoTMInput{random_rubble_masonry_rounded_stone_wall.jpg} &
\iTwoTMResult{chord/random_rubble_masonry_rounded_stone_wall_frame063_src062.jpg} &
\iTwoTMResult{matfusion/random_rubble_masonry_rounded_stone_wall_frame063_src062.jpg} &
\iTwoTMResult{Our_full/random_rubble_masonry_rounded_stone_wall_frame063_src248.jpg} &
\iTwoTMResult{GT/random_rubble_masonry_rounded_stone_wall_frame063_src062.jpg} &
\iTwoTMResult{LRM/random_rubble_masonry_rounded_stone_wall_frame063_src248.jpg} \\
\end{tabular}
\captionof{figure}{Comparison on material generation from a single input photograph. The input image is shown in the first column. The outputs are rendered from an oblique viewpoint, significantly different from the input, to better evaluate reconstruction quality. ``Our full'' denotes the complete pipeline: image-conditioned video generation followed by LRM reconstruction. ``Our LRM'' applies the LRM directly to synthetic rendered videos, isolating the reconstruction quality from the video generation step. Additional examples are shown in supplementary.}
\label{fig:i2t_comparison}
\end{minipage}%
\hfill
\begin{minipage}[t]{0.26\textwidth}
\centering
{\small
\setlength{\tabcolsep}{2.5pt}
\begin{tabular}{lccc}
\toprule
Method & MSE$\downarrow$ & LPIPS$\downarrow$ & PSNR$\uparrow$ \\
\midrule
\multicolumn{4}{@{}l}{\textit{Displayed frame}} \\
  CHORD     & 0.0469 & 0.465 & 13.29 \\
  MatFusion & 0.0271 & 0.594 & 15.67 \\
  Ours      & \textbf{0.0142} & \textbf{0.292} & \textbf{18.47} \\
\midrule
\multicolumn{4}{@{}l}{\textit{Test views}} \\
  CHORD     & 0.0332 & 0.400 & 14.80 \\
  MatFusion & 0.0254 & 0.572 & 15.96 \\
  Ours      & \textbf{0.0105} & \textbf{0.260} & \textbf{19.81} \\
\bottomrule
\end{tabular}%
}
\captionof{table}{Quantitative comparison on the image-to-texture test materials. ``Displayed frame'' evaluates the frame shown in the figure; ``Test views'' evaluates additional held-out views.}
\label{tab:i2t_comparison}
\end{minipage}
\end{table*}

\begin{figure}[tb]
    \centering
    \setlength{\tabcolsep}{1pt}
    
    \def\cropTop{140}
    
    \resizebox{\columnwidth}{!}{%
    \begin{tabular}{ccccc}
      & \multicolumn{2}{c}{Rendering} & \multicolumn{2}{c}{Offset} \\
      \cmidrule(lr){2-3} \cmidrule(lr){4-5}
      & Direct Optim. & LRM & Direct Optim. & LRM \\[2pt]
  
      \multirow{2}{*}[0.8em]{\rotatebox{90}{Synthetic}} &
      \includegraphics[width=2.5cm, trim=0 0 0 \cropTop, clip]{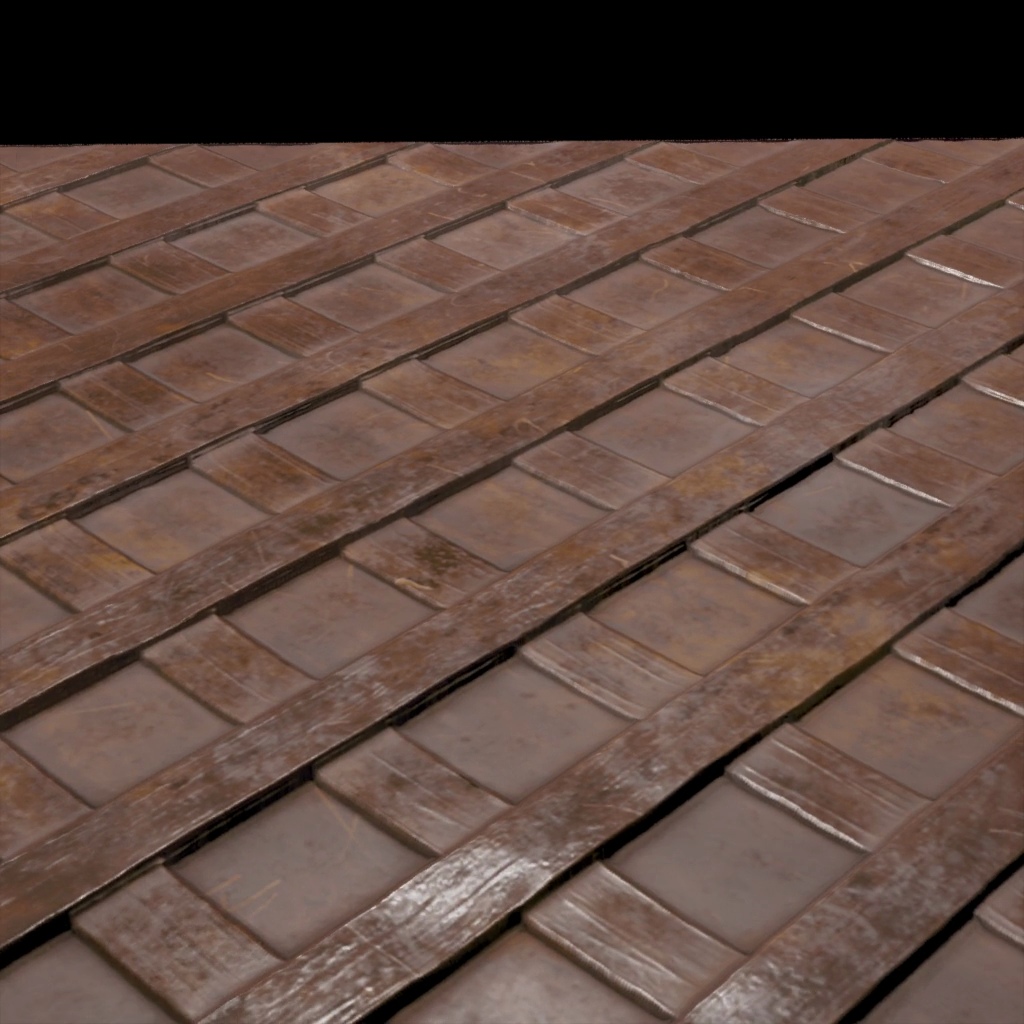} &
      \includegraphics[width=2.5cm, trim=0 0 0 \cropTop, clip]{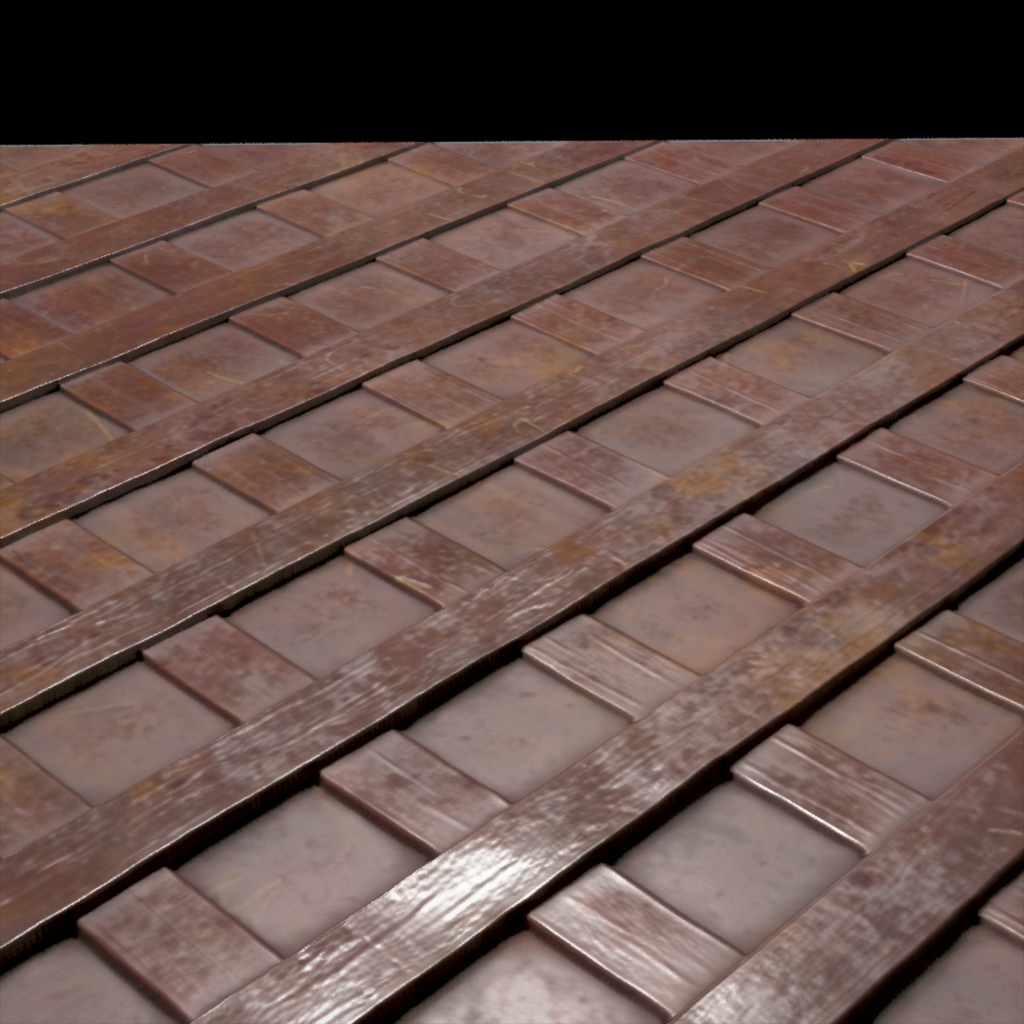} &
      \includegraphics[width=2.5cm, trim=0 0 0 \cropTop, clip]{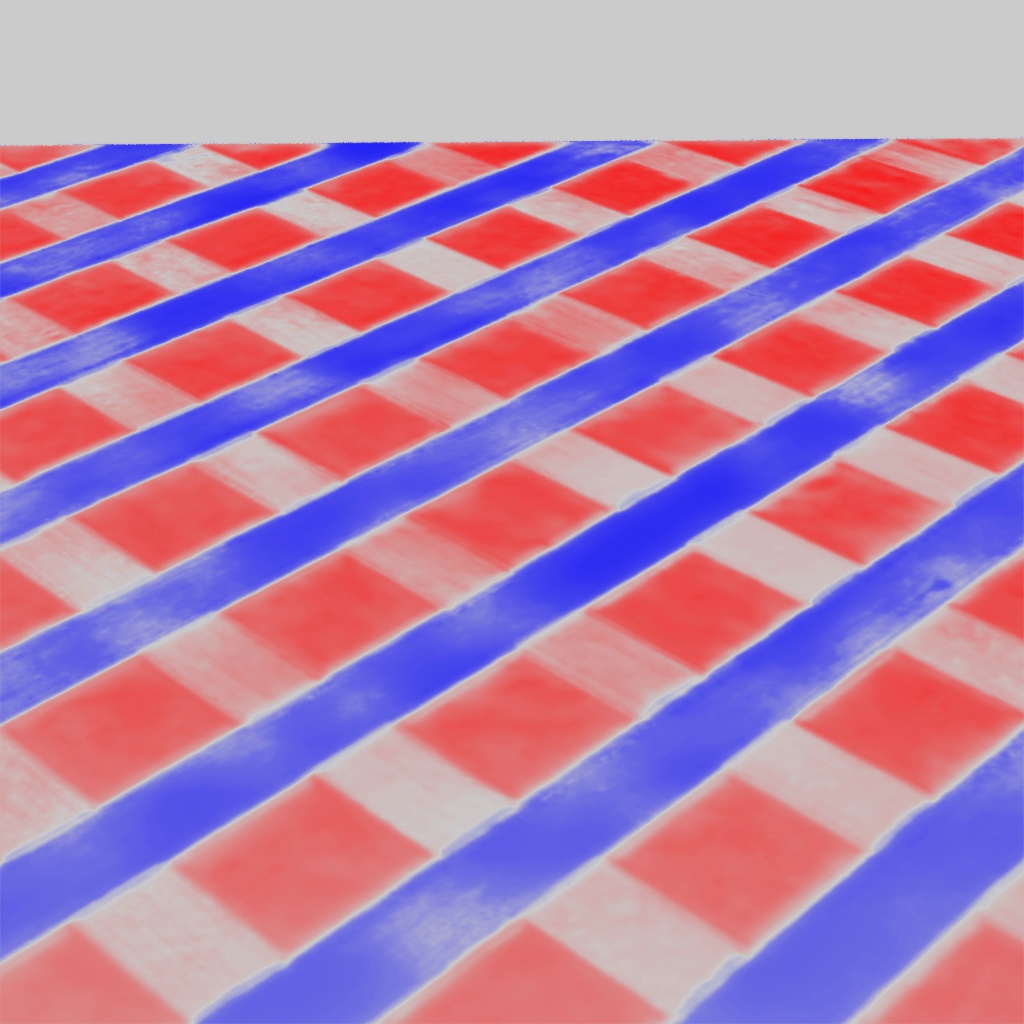} &
      \includegraphics[width=2.5cm, trim=0 0 0 \cropTop, clip]{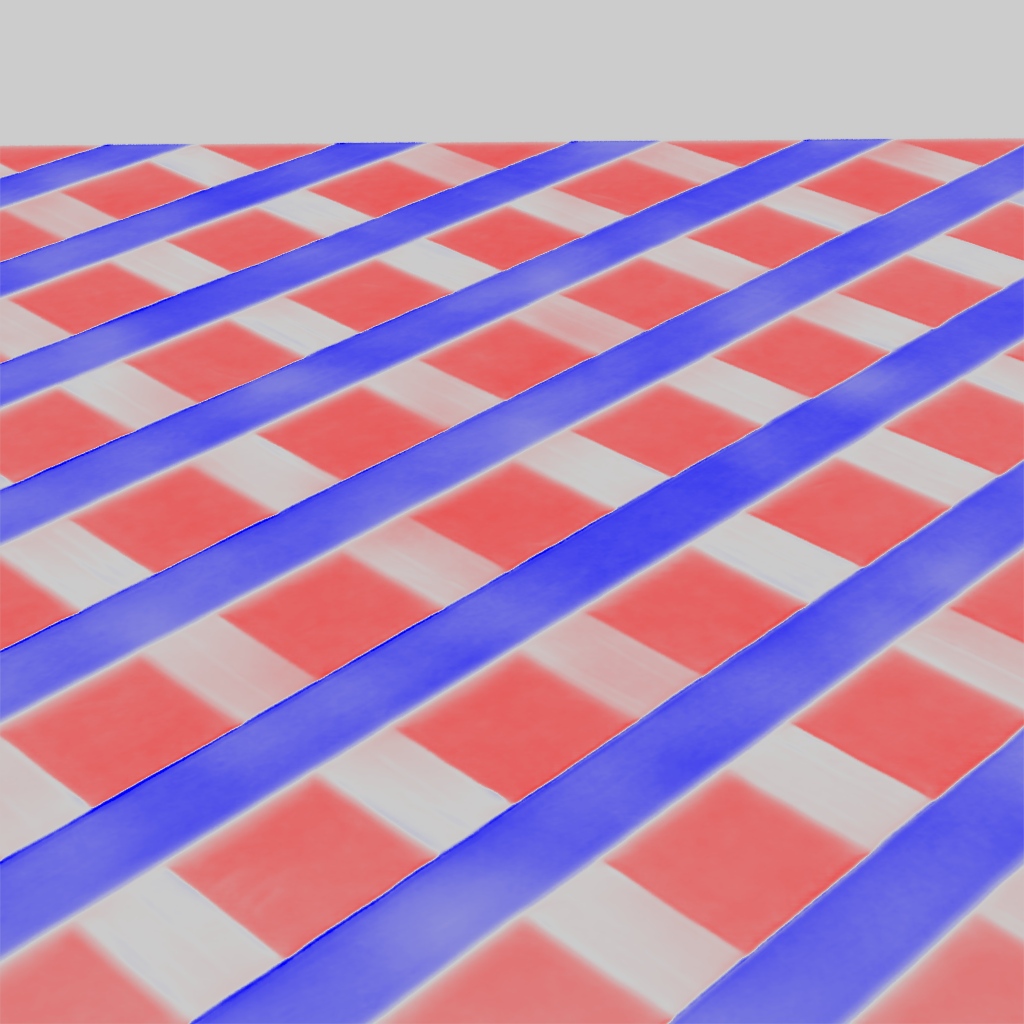} \\
      &
      \includegraphics[width=2.5cm, trim=0 0 0 \cropTop, clip]{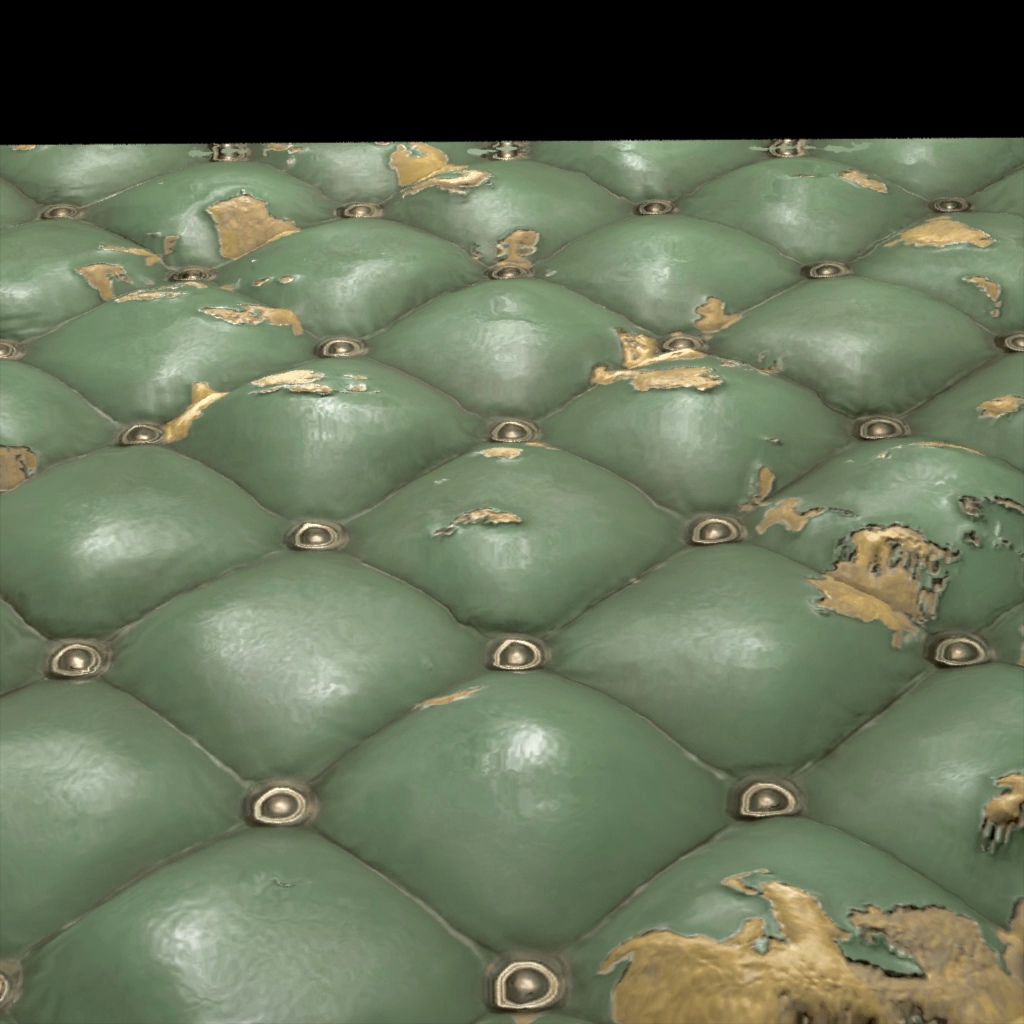} &
      \includegraphics[width=2.5cm, trim=0 0 0 \cropTop, clip]{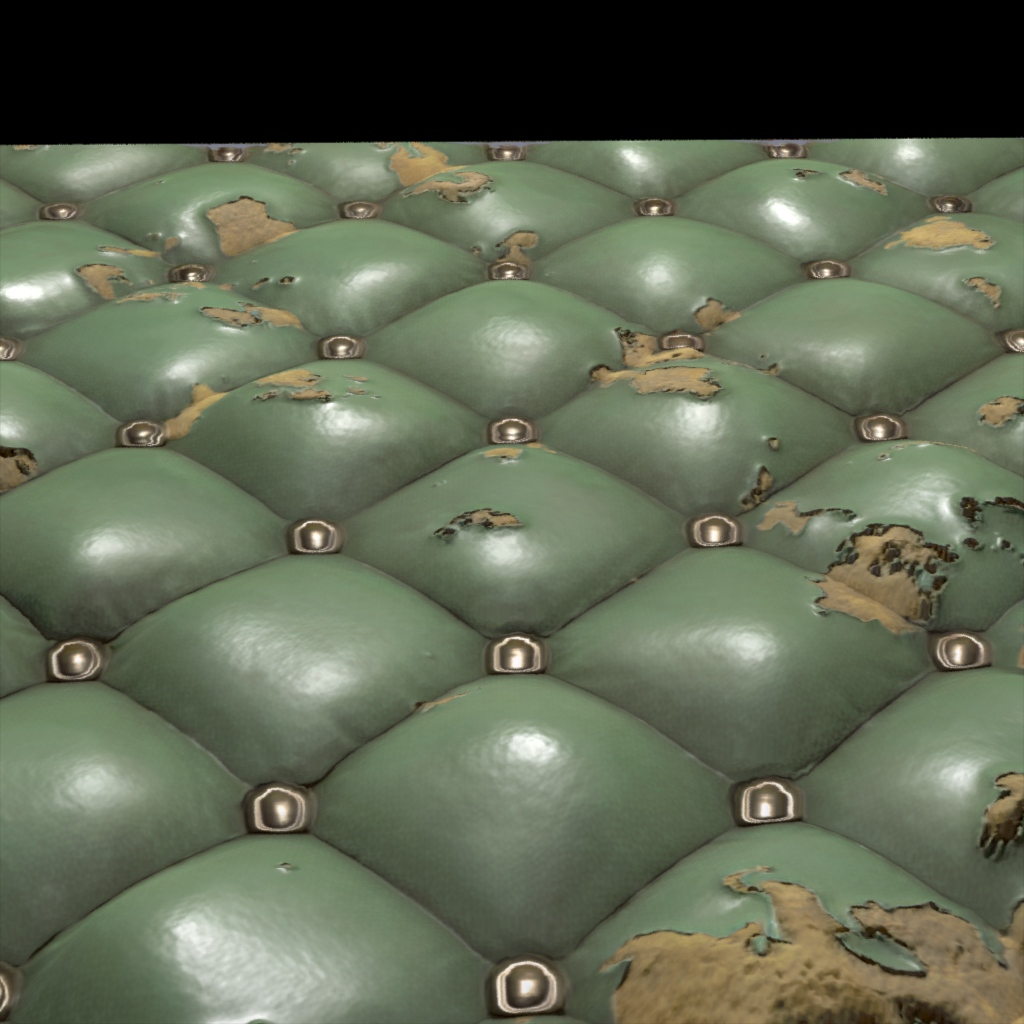} &
      \includegraphics[width=2.5cm, trim=0 0 0 \cropTop, clip]{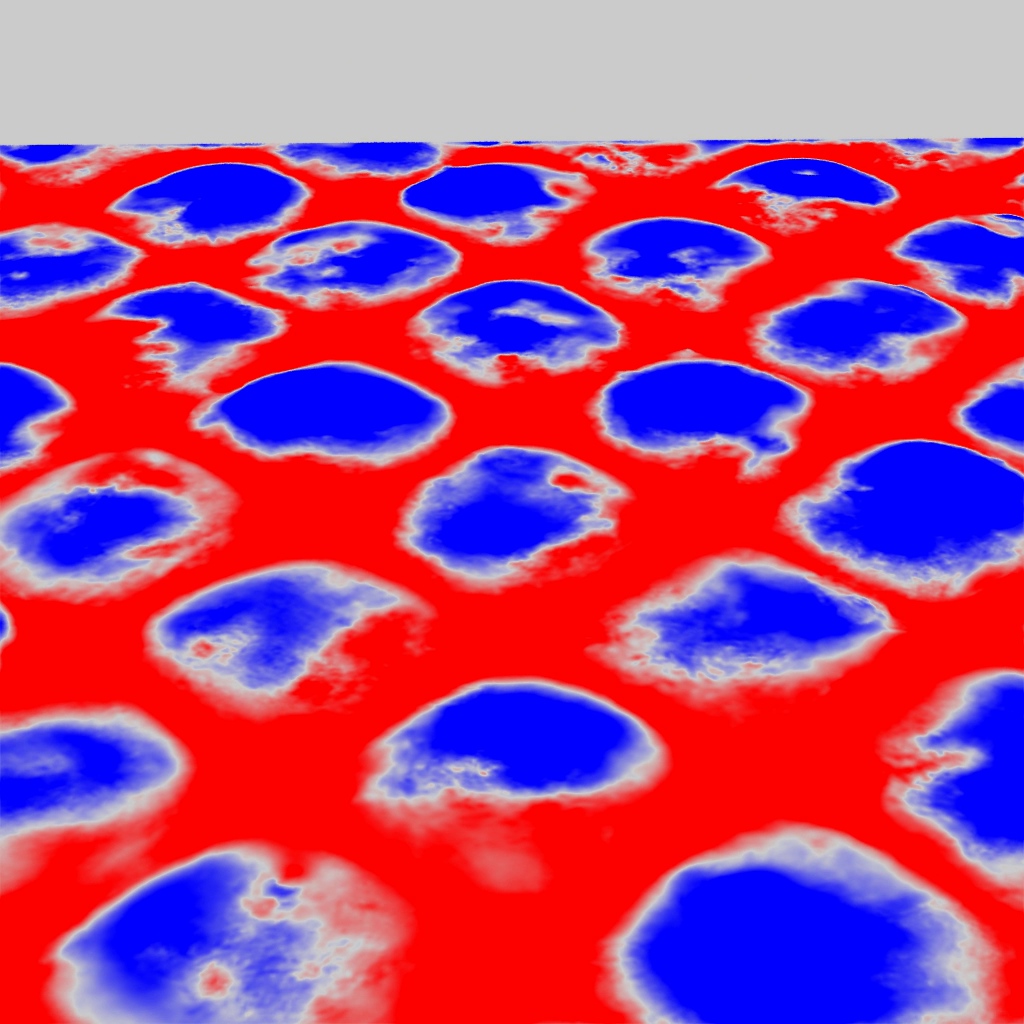} &
      \includegraphics[width=2.5cm, trim=0 0 0 \cropTop, clip]{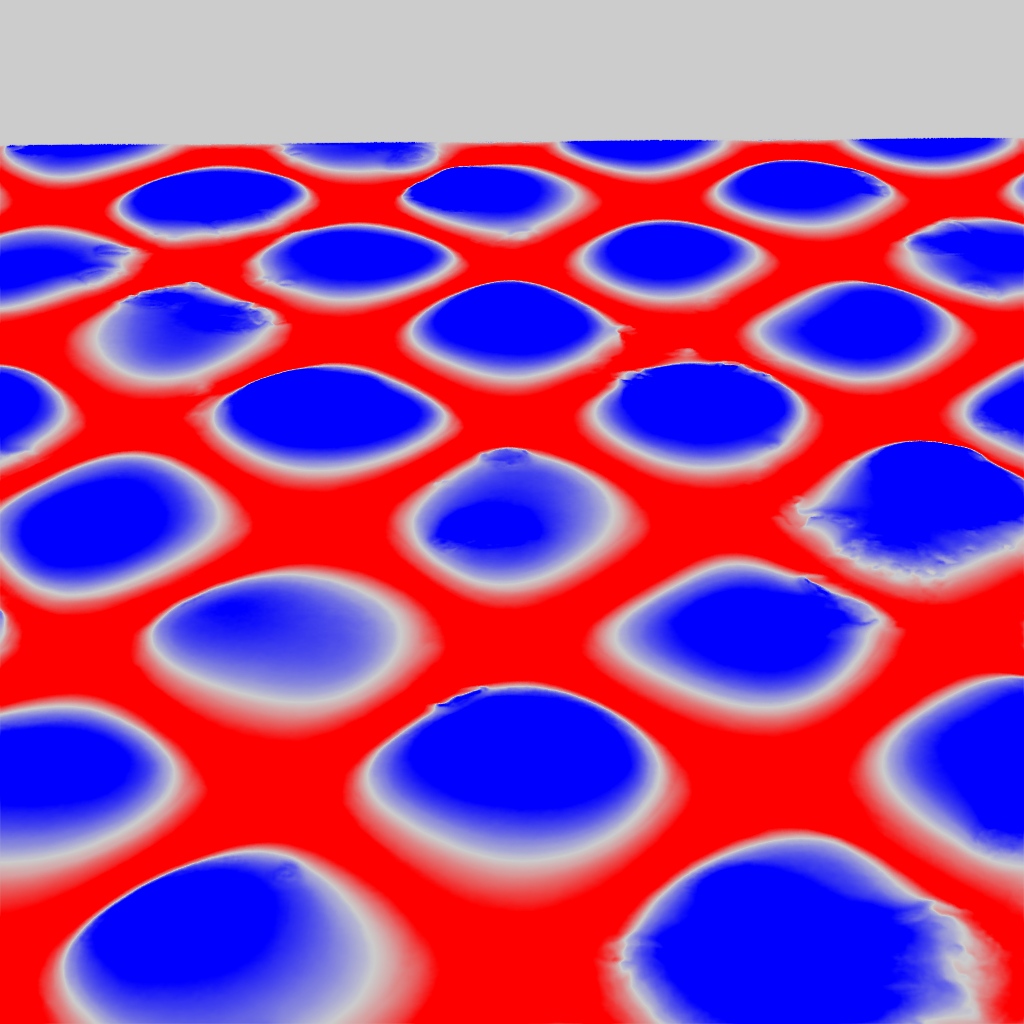} \\[2pt]
  
      \multirow{2}{*}[0.8em]{\rotatebox{90}{Diffusion}} &
      \includegraphics[width=2.5cm, trim=0 0 0 \cropTop, clip]{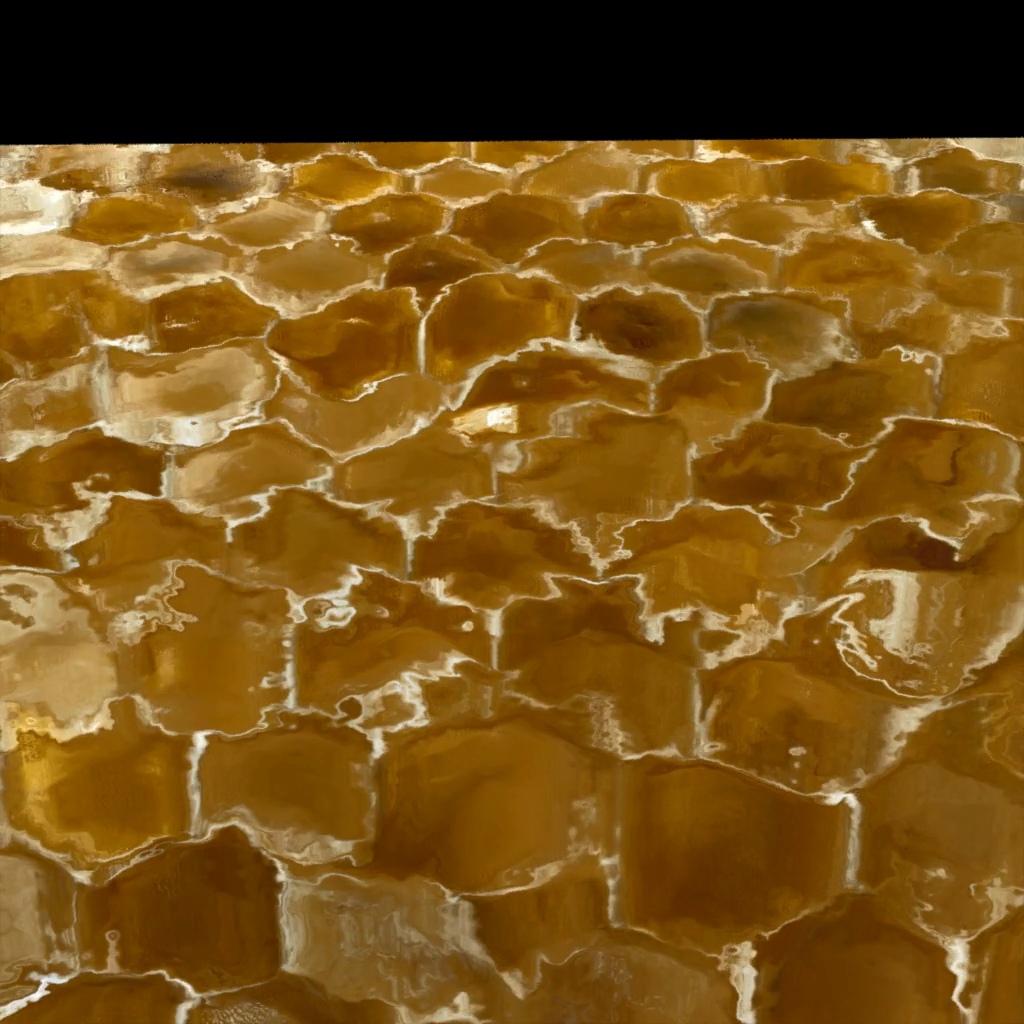} &
      \includegraphics[width=2.5cm, trim=0 0 0 \cropTop, clip]{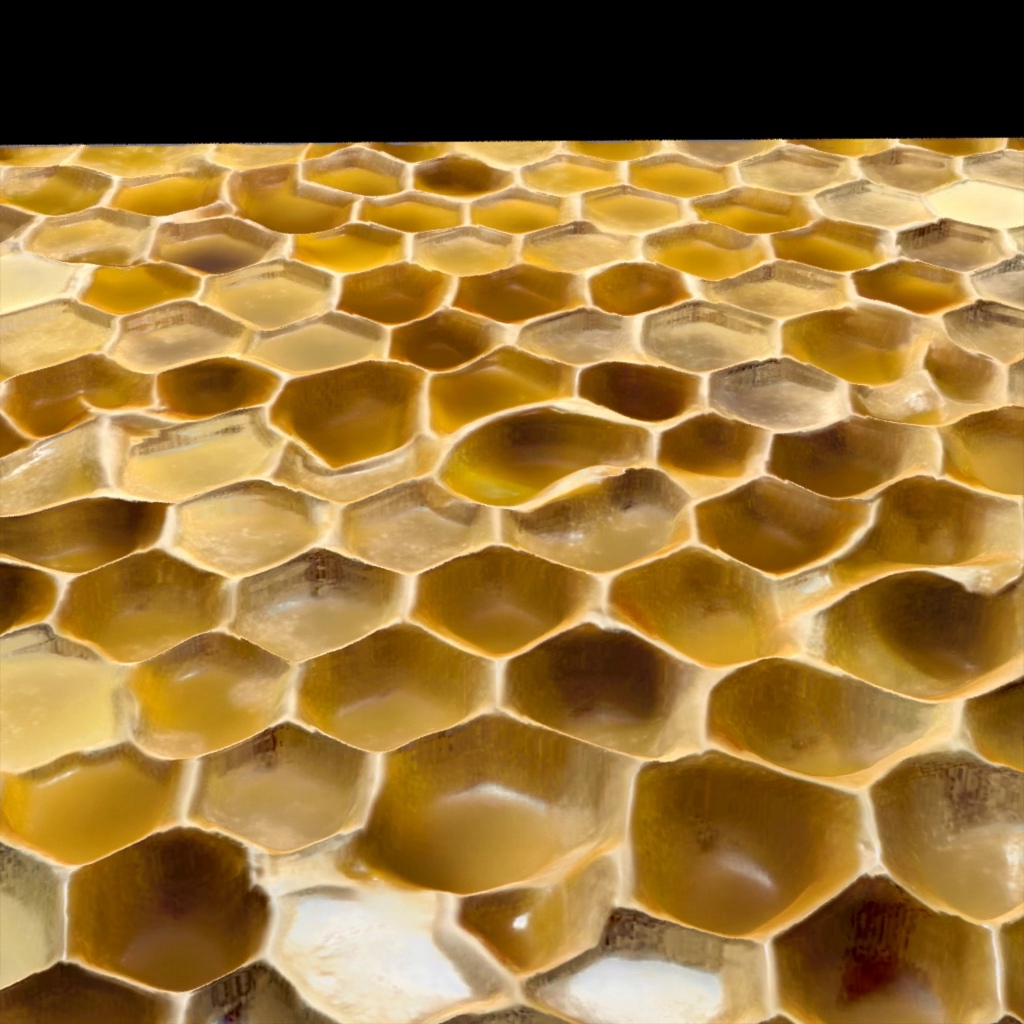} &
      \includegraphics[width=2.5cm, trim=0 0 0 \cropTop, clip]{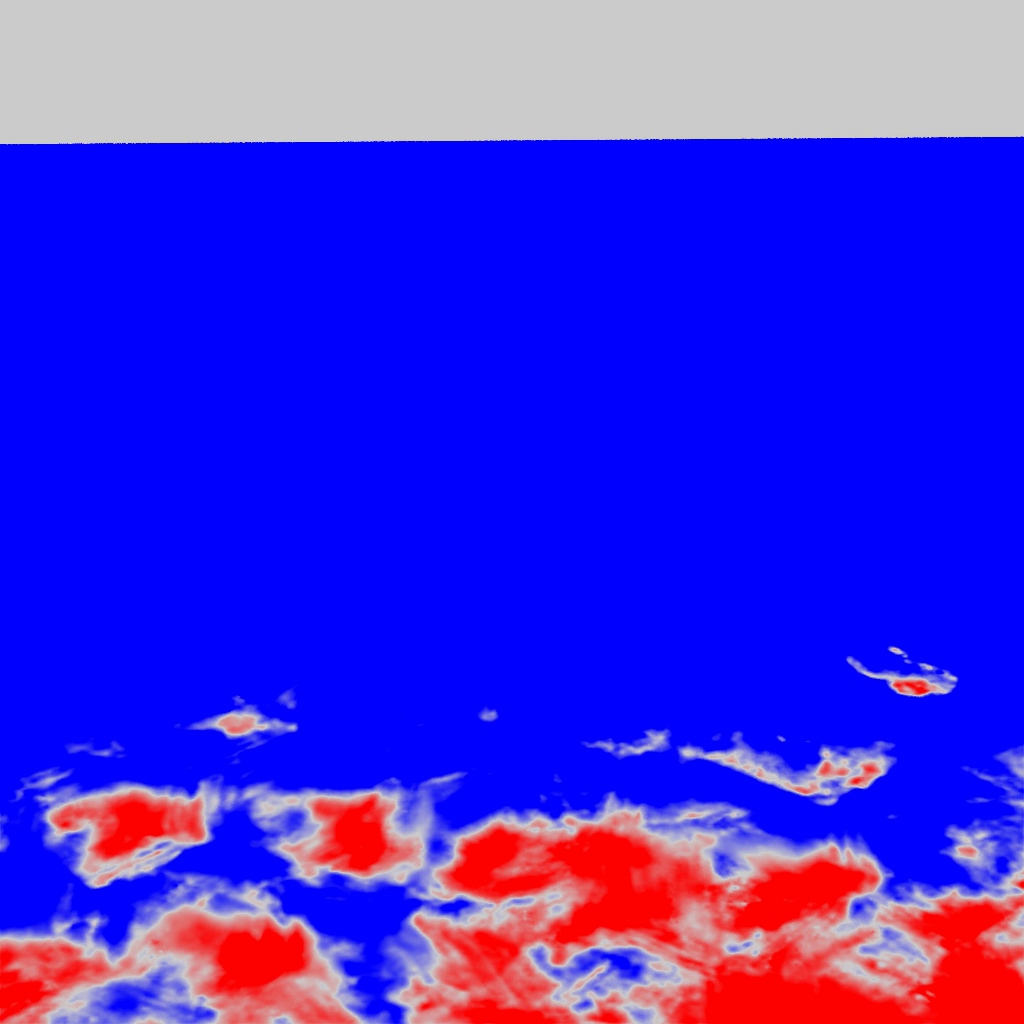} &
      \includegraphics[width=2.5cm, trim=0 0 0 \cropTop, clip]{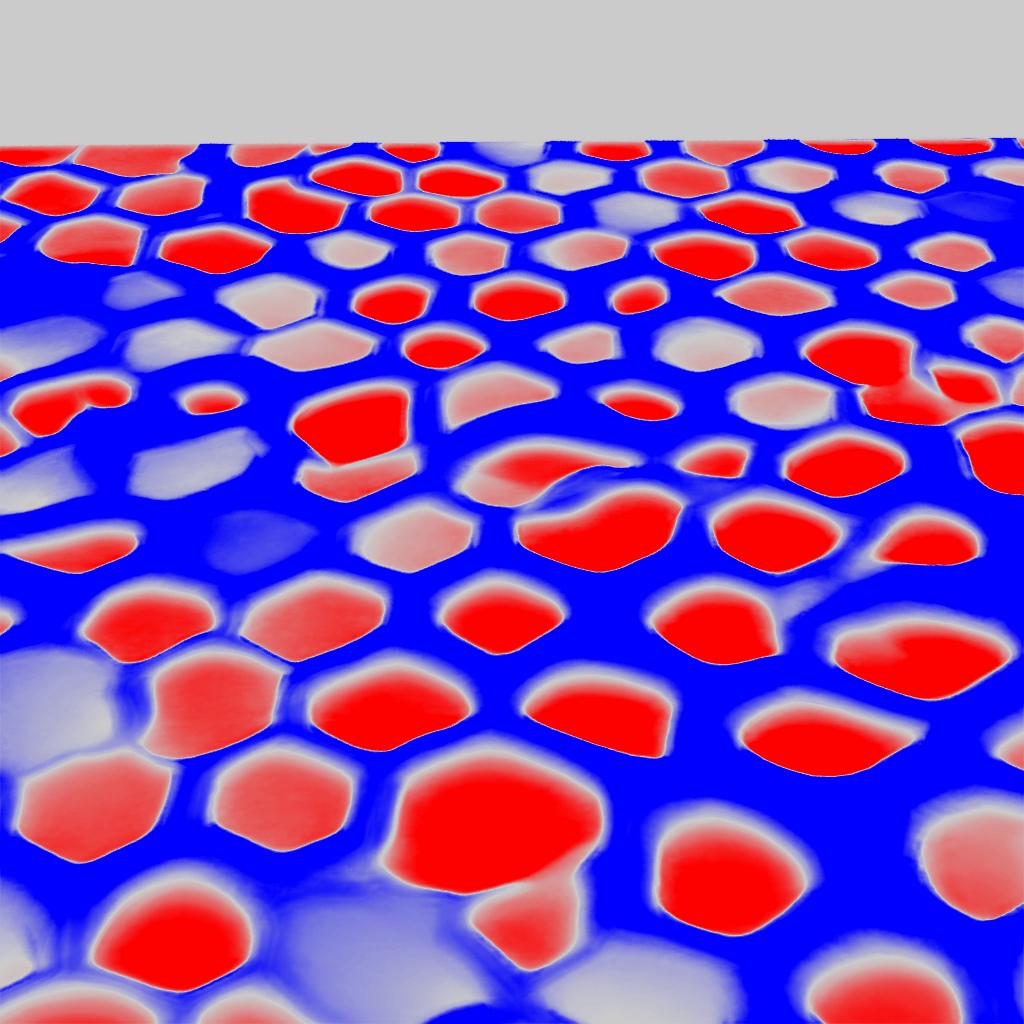} \\
      &
      \includegraphics[width=2.5cm, trim=0 0 0 \cropTop, clip]{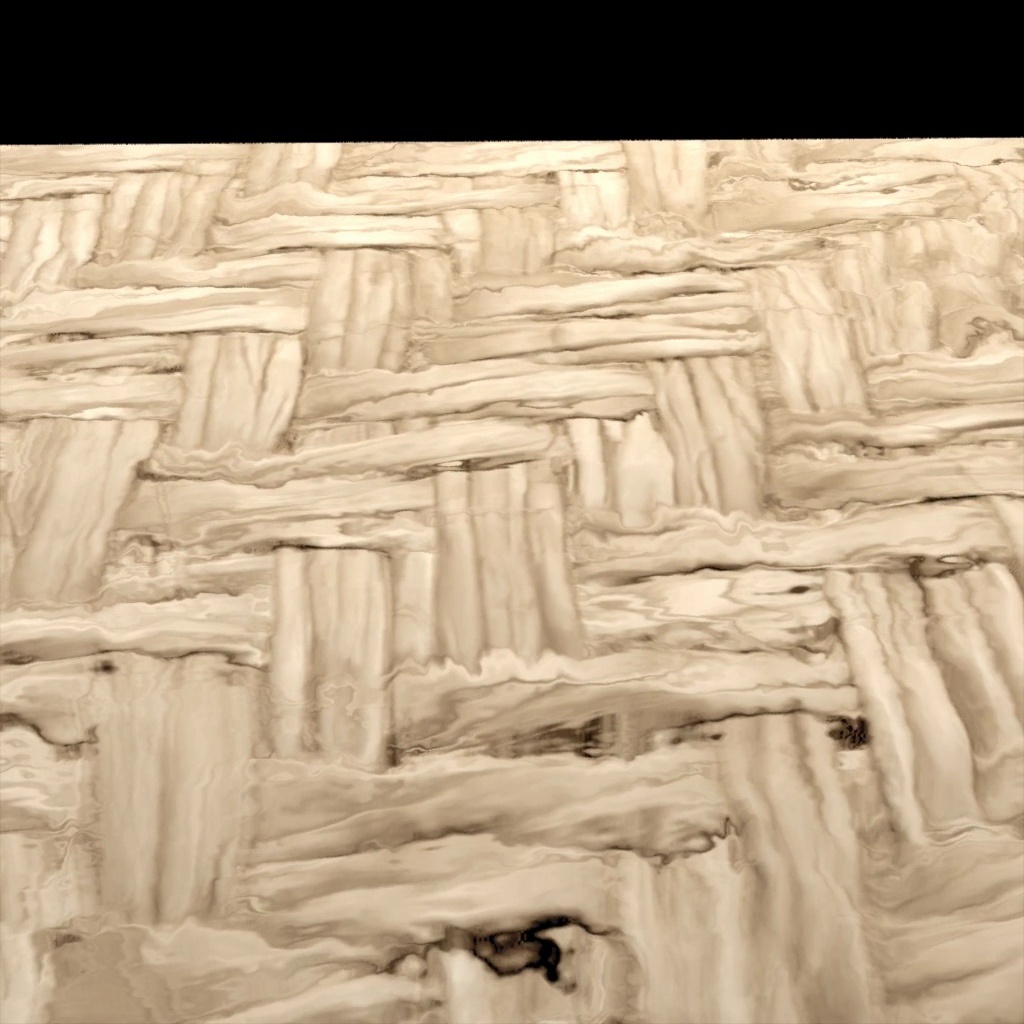} &
      \includegraphics[width=2.5cm, trim=0 0 0 \cropTop, clip]{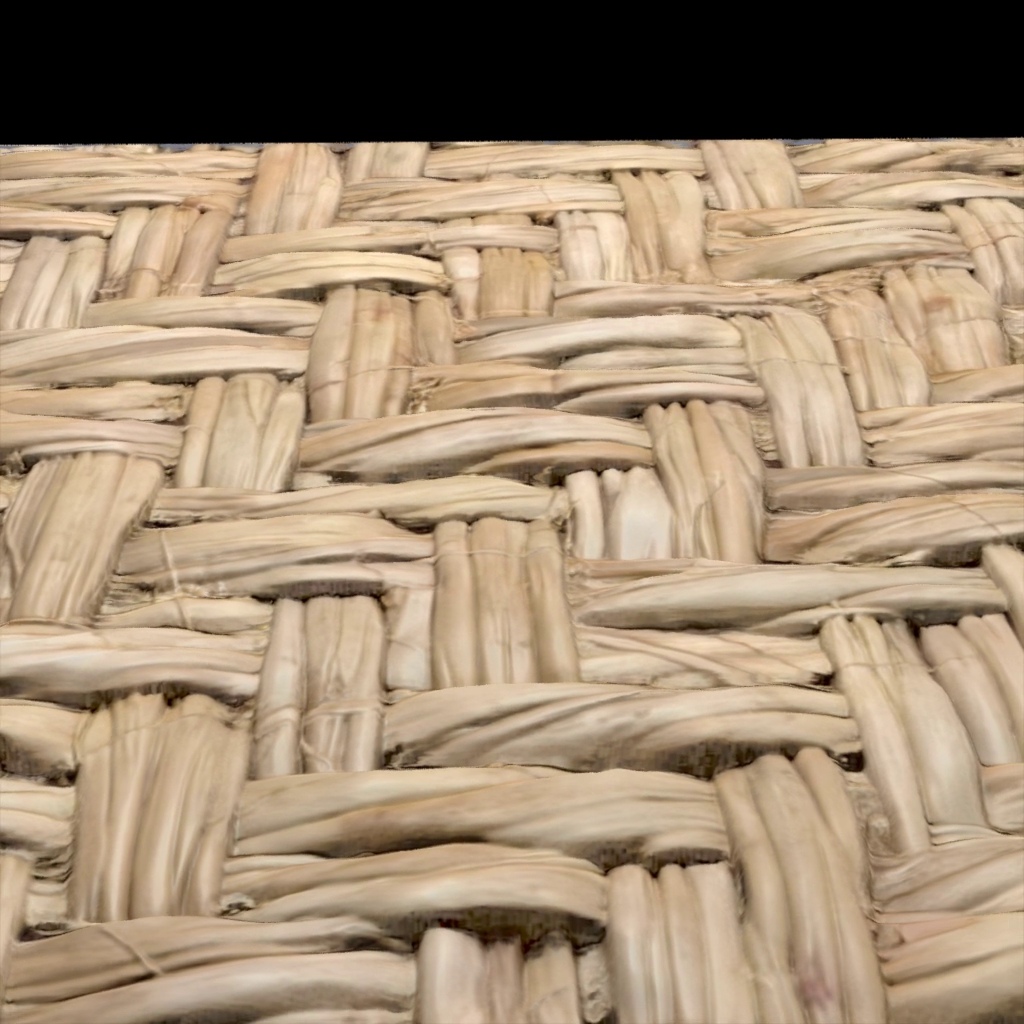} &
      \includegraphics[width=2.5cm, trim=0 0 0 \cropTop, clip]{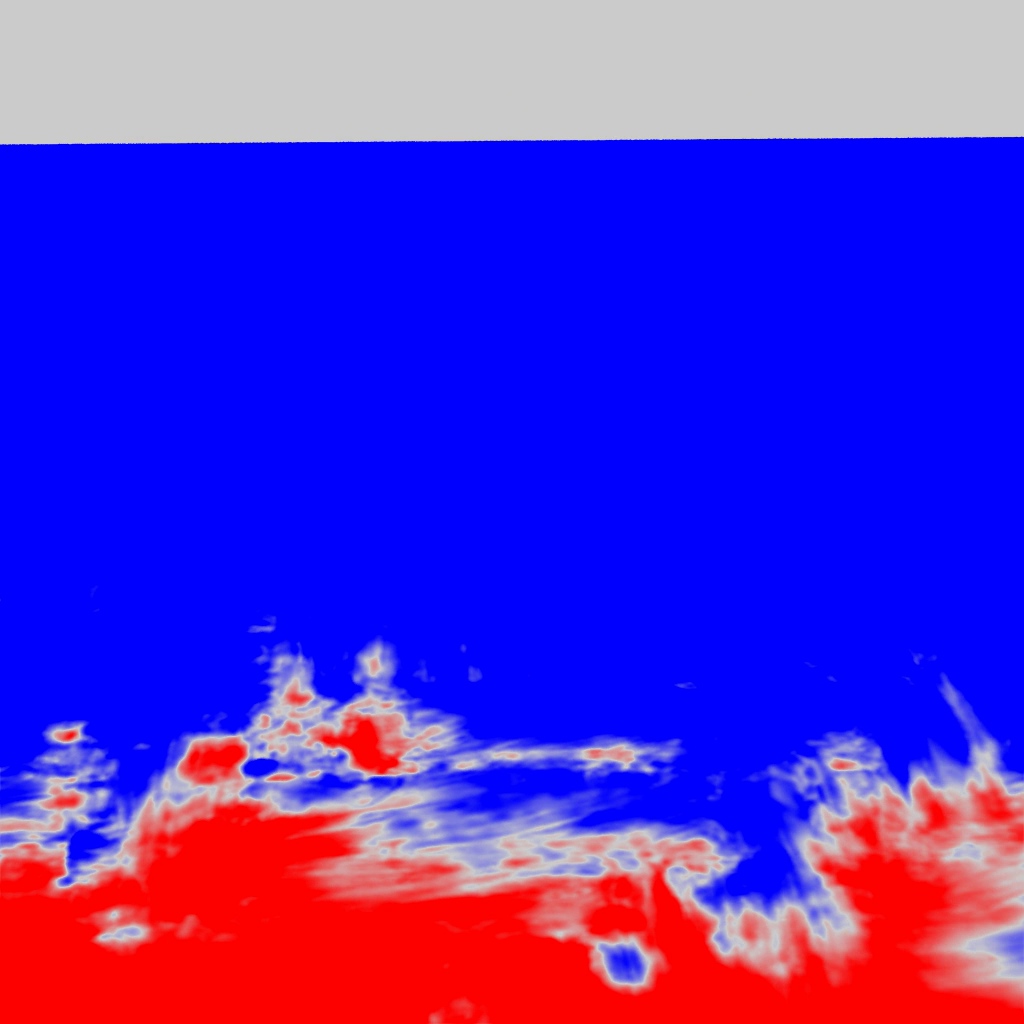} &
      \includegraphics[width=2.5cm, trim=0 0 0 \cropTop, clip]{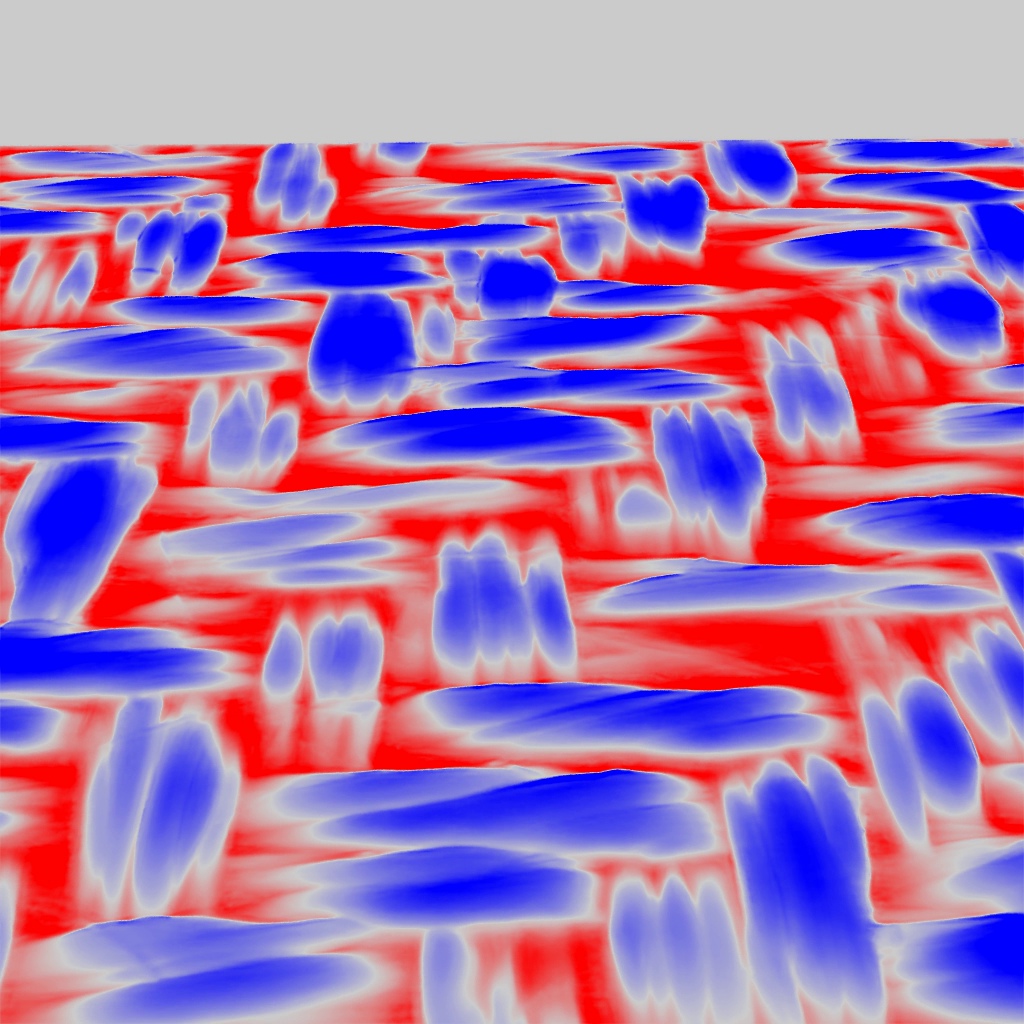} \\
    \end{tabular}}
    
   \caption{We compare direct optimization with LRM by showing renderings of unseen views during training. Direct optimization fails to reach the correct offset in diffusion materials, likely due to the misalignments in the generated videos. Even for synthetic materials with no misalignments, the LRM has better generalization to unseen views, since it was trained on more camera/light settings. Full videos are included in the supplementary video.}
    \label{fig:direct_optim}
  \end{figure}

\paragraph{Quantitative evaluation.}
A key hypothesis of our work is that finetuning the video model on synthetic, measurement-like trajectories can steer the generator toward our protocol \emph{without} collapsing to a synthetic-rendered regime, i.e., it retains the video model's realistic prior.
We therefore evaluate two complementary signals (\autoref{evaluation}a).
We compute FVD to OpenVid-1M~\cite{nan2024openvid} to measure distributional closeness to real videos in a standard video embedding space.
The FVD serves as an indication of whether finetuning preserves real-video statistics rather than drifting toward synthetic-rendered artifacts. Our generated clips (783 samples) achieve a lower FVD (118.04) than MatSynth renderings (138.86).
We additionally report RealMatScore~\cite{realmat}, a material-focused perceptual metric, where our results improve over MatSynth (0.6289 vs.\ 0.5463).
These results suggest that finetuning retains the pretrained model's real-video appearance prior, rather than collapsing toward a purely synthetic-rendered regime, as further supported by the nearest-neighbor analysis in the supplementary material.

\section{Ablation Study}



\paragraph{Perspective vs.\ rectified projection (\autoref{evaluation}b).}
\label{sec:ablation_trajectory}
Traditional BTF acquisition employs orthographic cameras with rectified 
sensing, ensuring pixel-to-surface correspondence across frames.
We compare this against our perspective projection approach by training 
two video models on the same material set, differing only in projection mode.
The perspective model preserves material realism at the cost of 
correspondence ambiguity, yet still achieves competitive reconstruction 
quality on generated videos.

\paragraph{Video model finetuning strategies (\autoref{evaluation}b).} We provide two finetuning variants with different resource-quality trade-offs. Our LoRA variant (rank 32) requires less GPU memory and training time while still producing realistic materials that follow the measurement trajectory. Our full finetuning variant updates all DiT parameters and achieves the best trajectory coherence (RPC 0.2106 vs.\ 0.3102) and perceptual realism (RealMatScore 0.6289 vs.\ 0.6190, though this difference is very small). User can choose based on their computational budget and quality requirements.

\begin{table}[t]
\caption{(a) Appearance Evaluation. FVD to OpenVid-1M measures distributional closeness to real videos and is used as a proxy for preserving the pretrained video prior (not as a direct material-correctness metric).
RealMatScore measures material-focused perceptual appearance. (b) Ablation study of training the video model compares finetuning strategies and projection modes.}
\label{evaluation}
\centering
\small
\setlength{\tabcolsep}{2pt}
\begin{minipage}[t]{0.40\linewidth}
\centering
\textbf{(a) Appearance}\\[2pt]
\begin{tabular}{lcc}
\toprule
     & \shortstack{\phantom{Ours}\\MatSynth} & \shortstack{\phantom{Ours}\\Ours} \\
\midrule
FVD$\downarrow$ & 138.86 & \textbf{118.04} \\
RealMat$\uparrow$ & 0.5463 & \textbf{0.6289} \\
\bottomrule
\end{tabular}
\end{minipage}
\hfill
\begin{minipage}[t]{0.55\linewidth}
\centering
\textbf{(b) Ablation}\\[2pt]
\begin{tabular}{lccc}
\toprule
     & \shortstack{Ours\\LoRA} & \shortstack{Ours\\Finetune} & \shortstack{Rectified\\Projection} \\
\midrule
RPC$\downarrow$ & 0.3102 & \textbf{0.2106} & 0.2146 \\
RealMat$\uparrow$ & 0.6190 & \textbf{0.6289} & 0.5992 \\
\bottomrule
\end{tabular}
\end{minipage}
\end{table}

\begin{figure*}[tbh]
    \centering
    \setlength{\tabcolsep}{1pt}
    \renewcommand{\arraystretch}{0.5}

    \resizebox{\textwidth}{!}{%
    \begin{tabular}{@{}r@{\hspace{2pt}}ccccccccc@{}}
        \rotatebox{90}{\hspace{3pt}\footnotesize{Input condition}} &
        \includegraphics[width=0.100\textwidth]{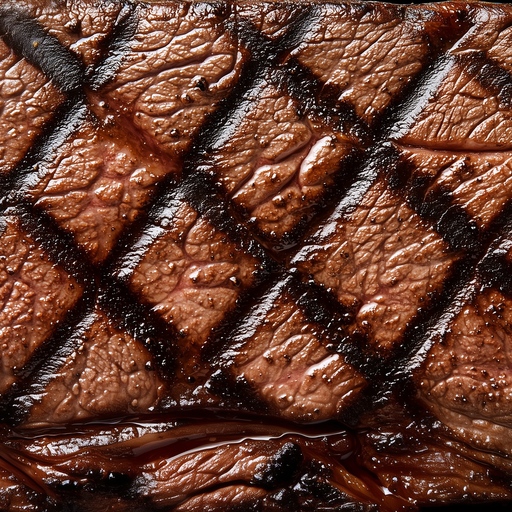} &
        \includegraphics[width=0.100\textwidth]{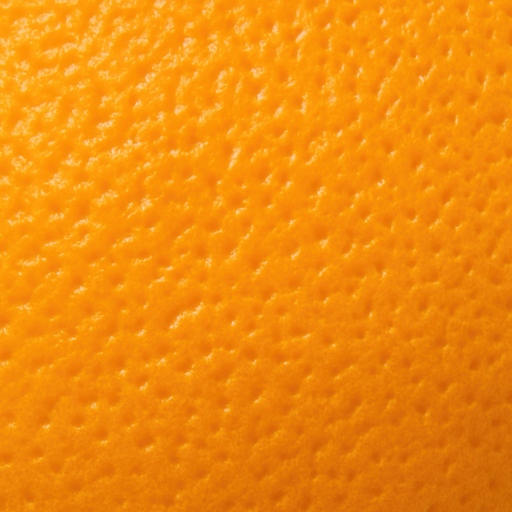} &
        \includegraphics[width=0.100\textwidth]{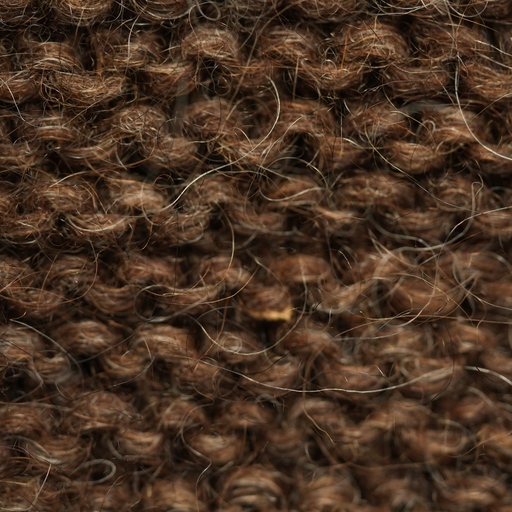} &
        \includegraphics[width=0.100\textwidth]{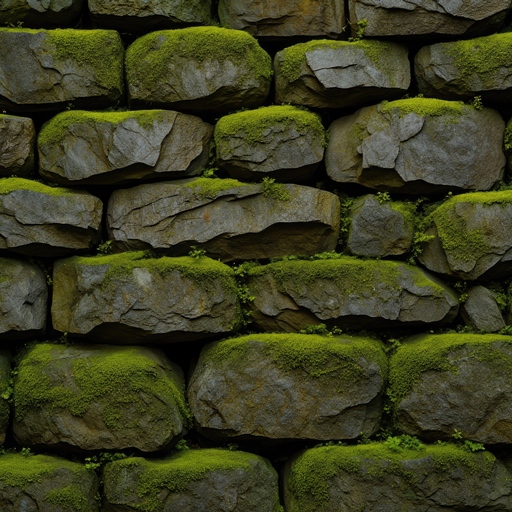} &
        \includegraphics[width=0.100\textwidth]{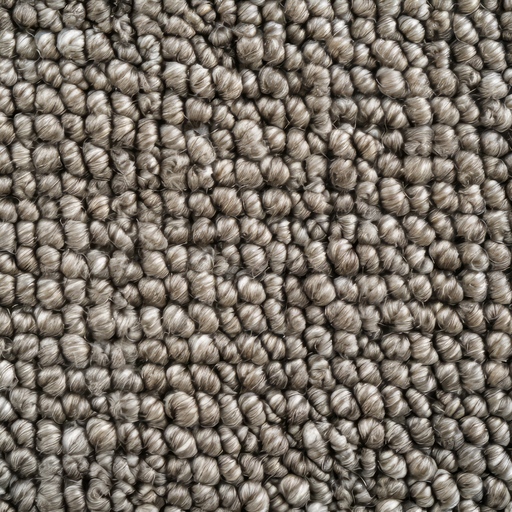} &
        \includegraphics[width=0.100\textwidth]{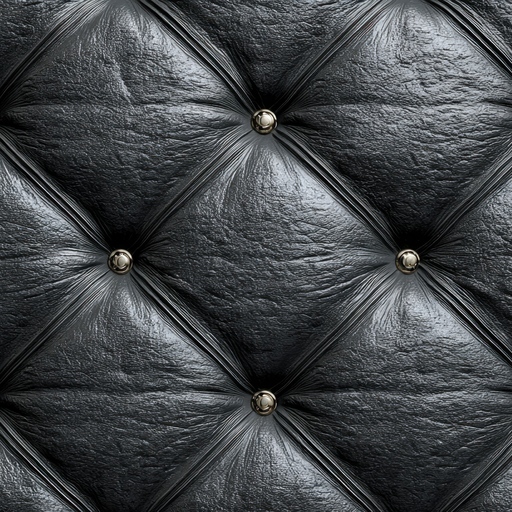} &
        \includegraphics[width=0.100\textwidth]{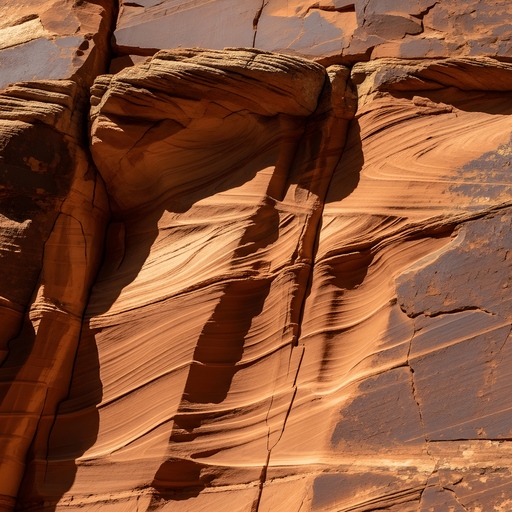} &
        \includegraphics[width=0.100\textwidth]{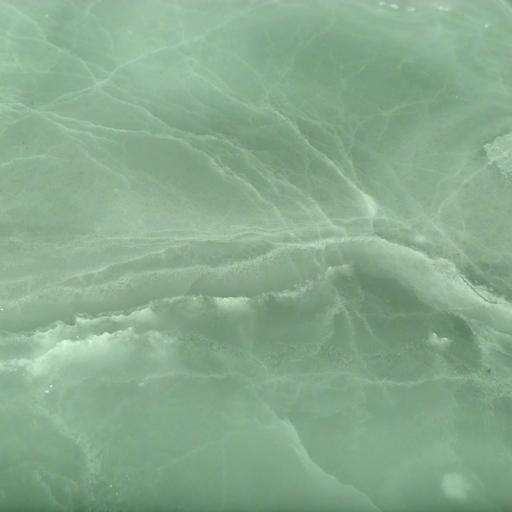} &
        \includegraphics[width=0.100\textwidth]{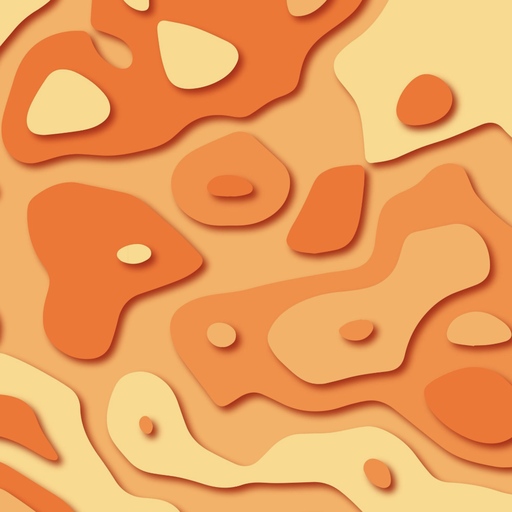} \\[1pt]
        \rotatebox{90}{\hspace{9pt}\footnotesize{Generated}} &
        \includegraphics[width=0.100\textwidth]{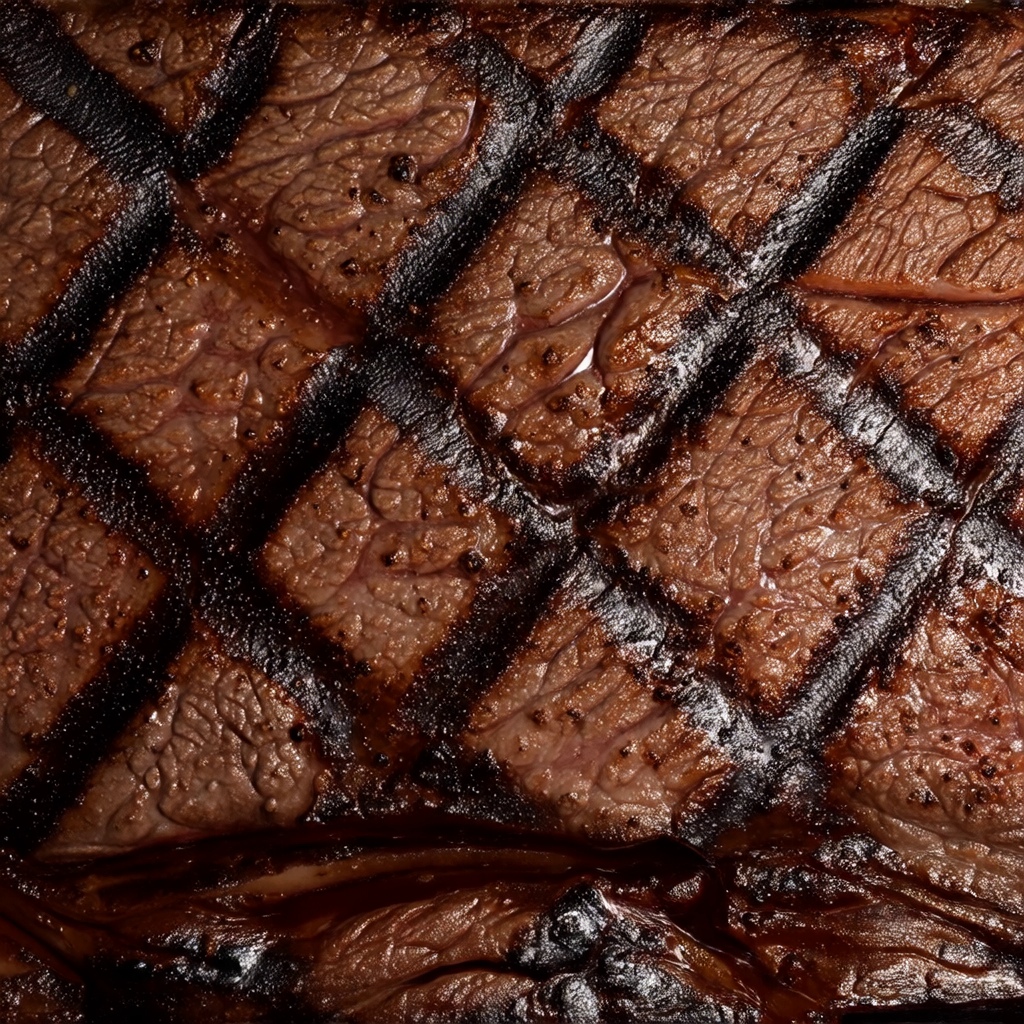} &
        \includegraphics[width=0.100\textwidth]{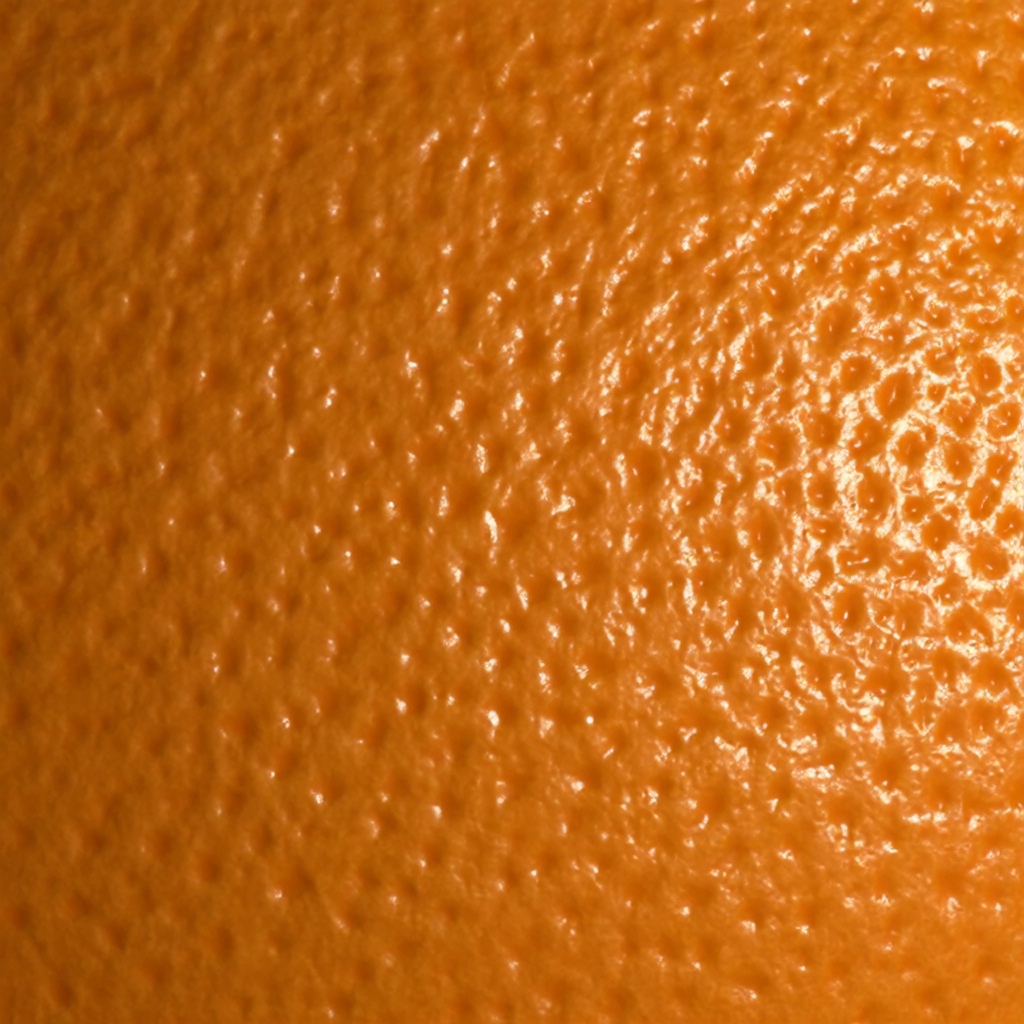} &
        \includegraphics[width=0.100\textwidth]{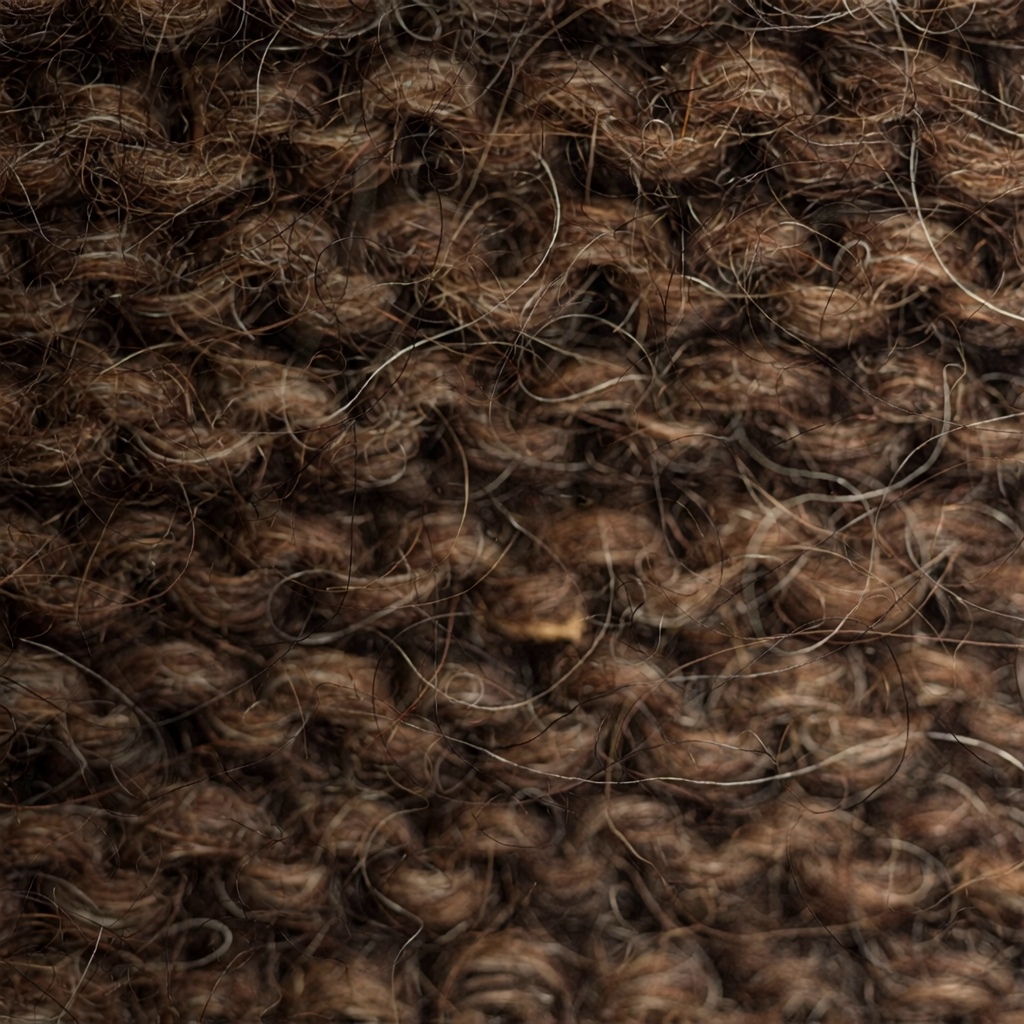} &
        \includegraphics[width=0.100\textwidth]{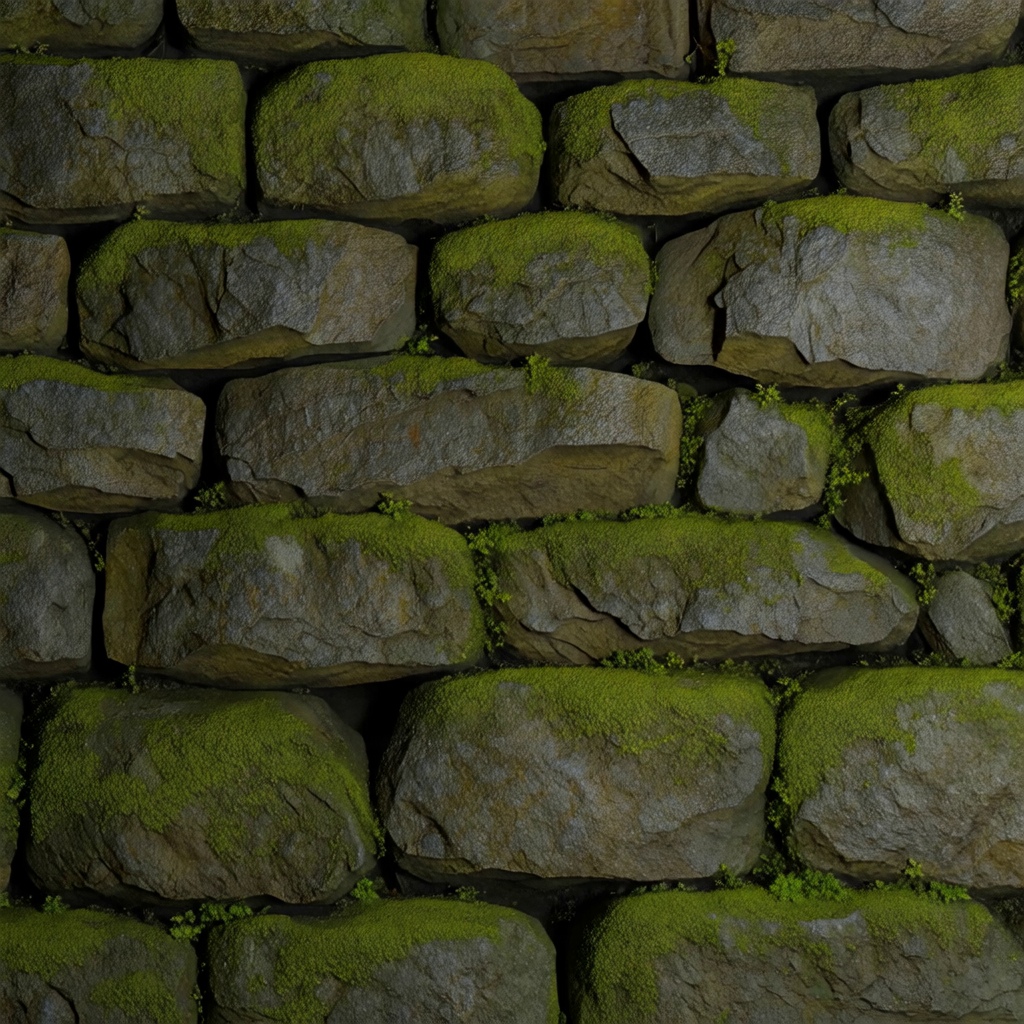} &
        \includegraphics[width=0.100\textwidth]{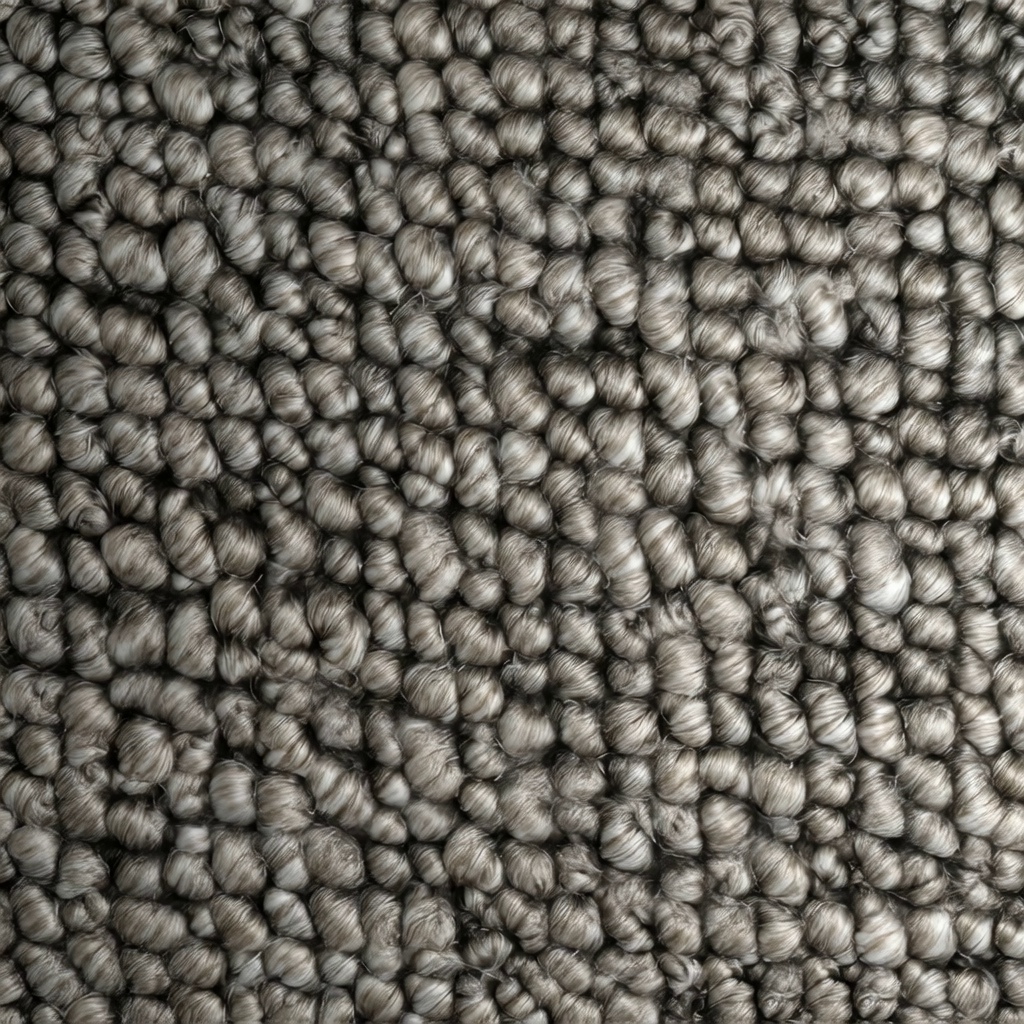} &
        \includegraphics[width=0.100\textwidth]{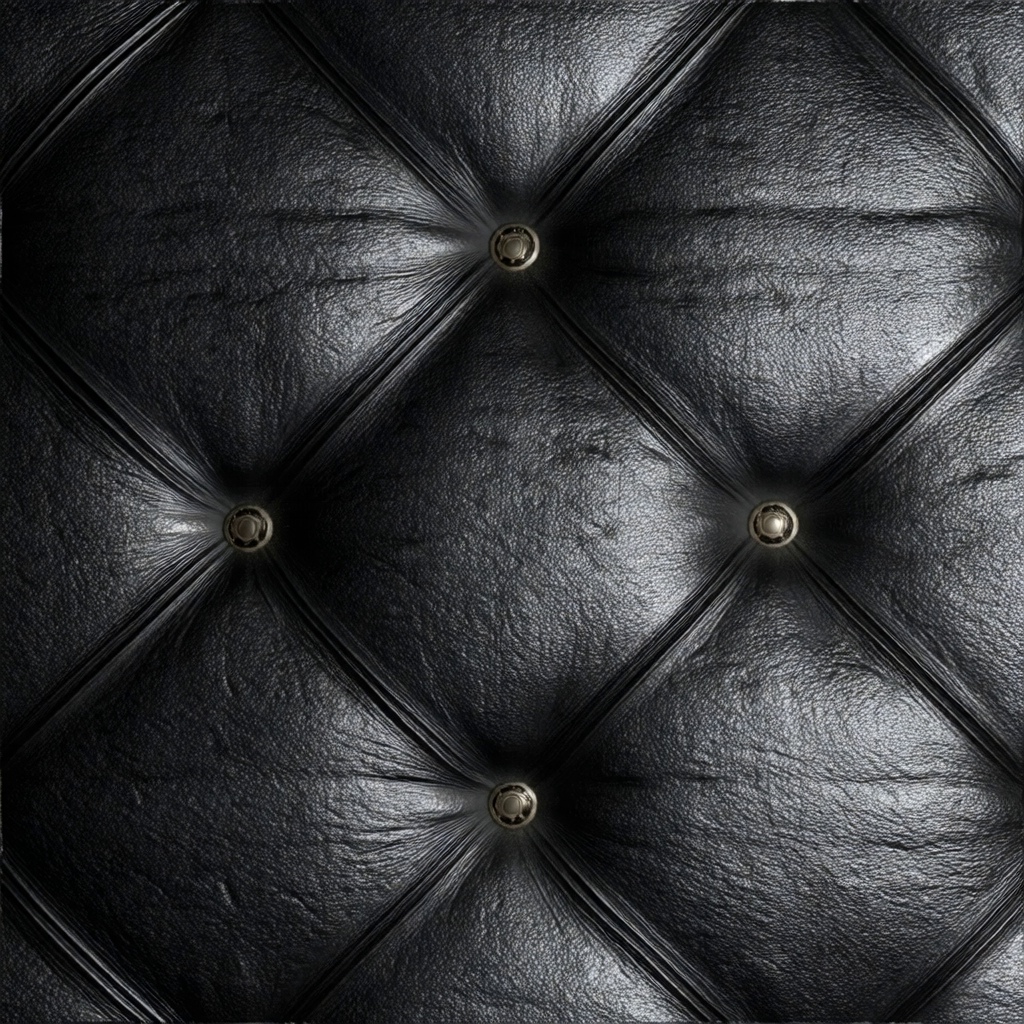} &
        \includegraphics[width=0.100\textwidth]{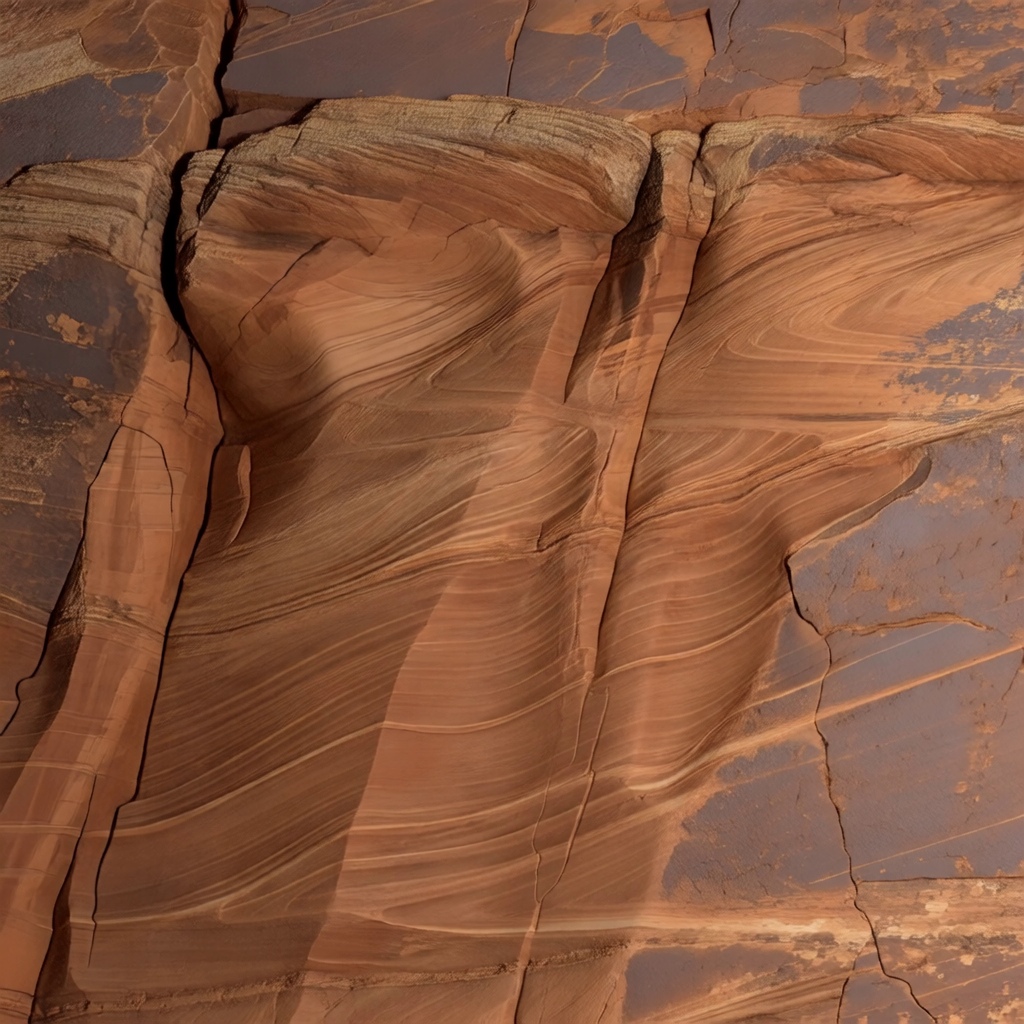} &
        \includegraphics[width=0.100\textwidth]{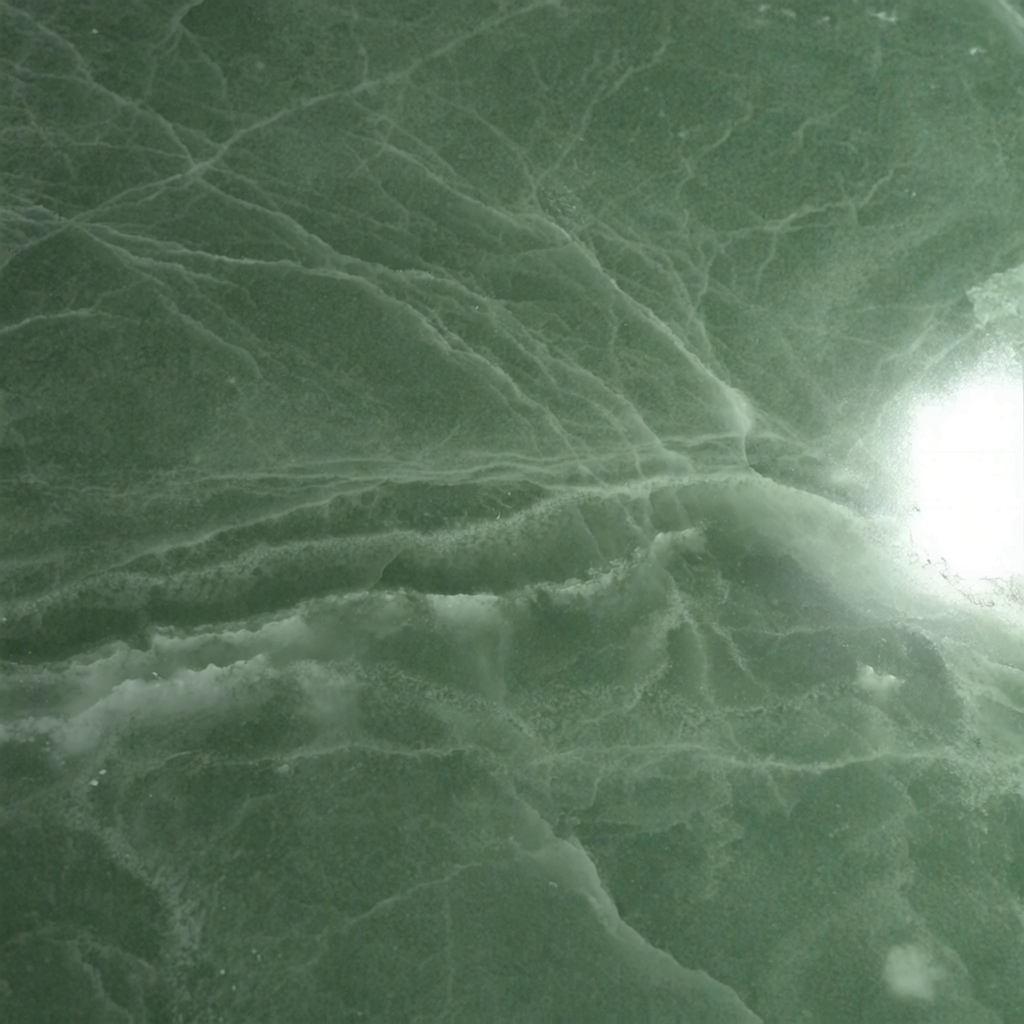} &
        \includegraphics[width=0.100\textwidth]{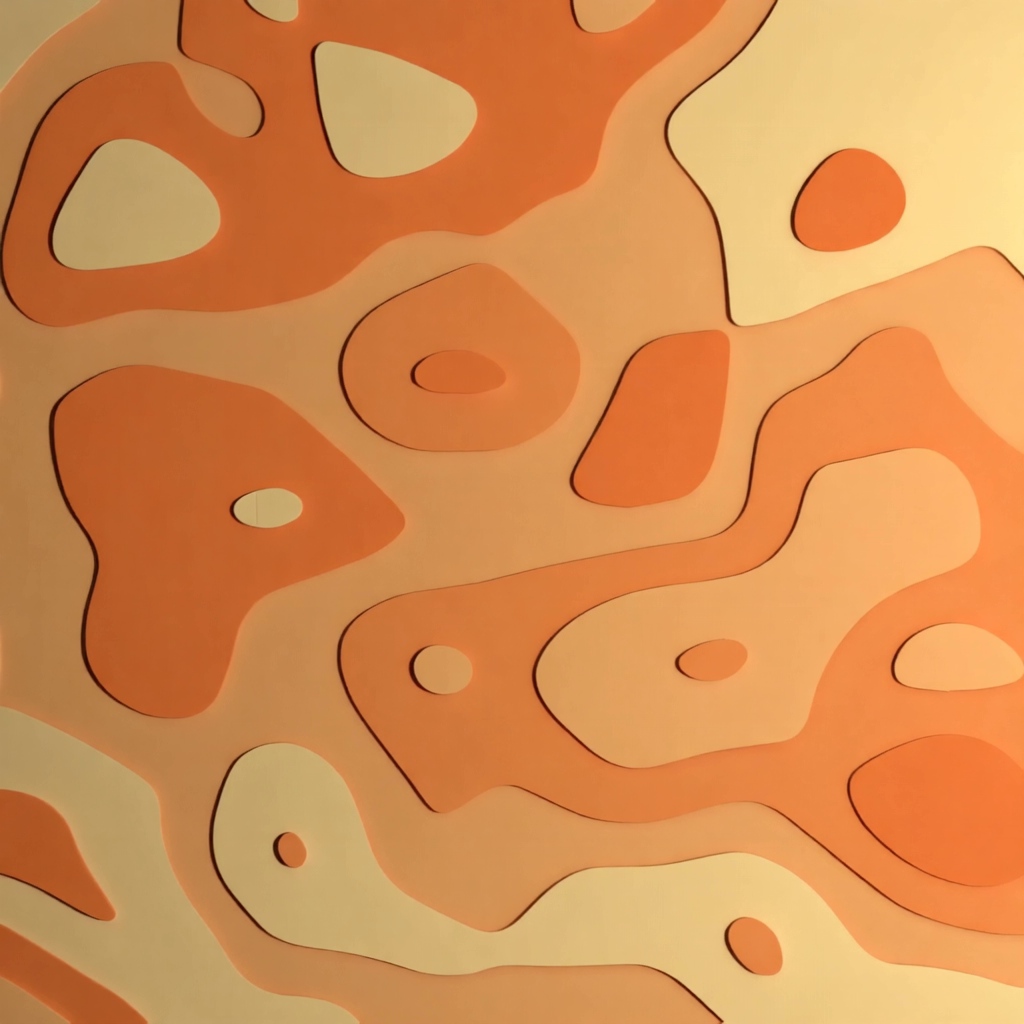} \\[1pt]
        \rotatebox{90}{\hspace{5pt}\footnotesize{Reconstructed}} &
        \includegraphics[width=0.100\textwidth]{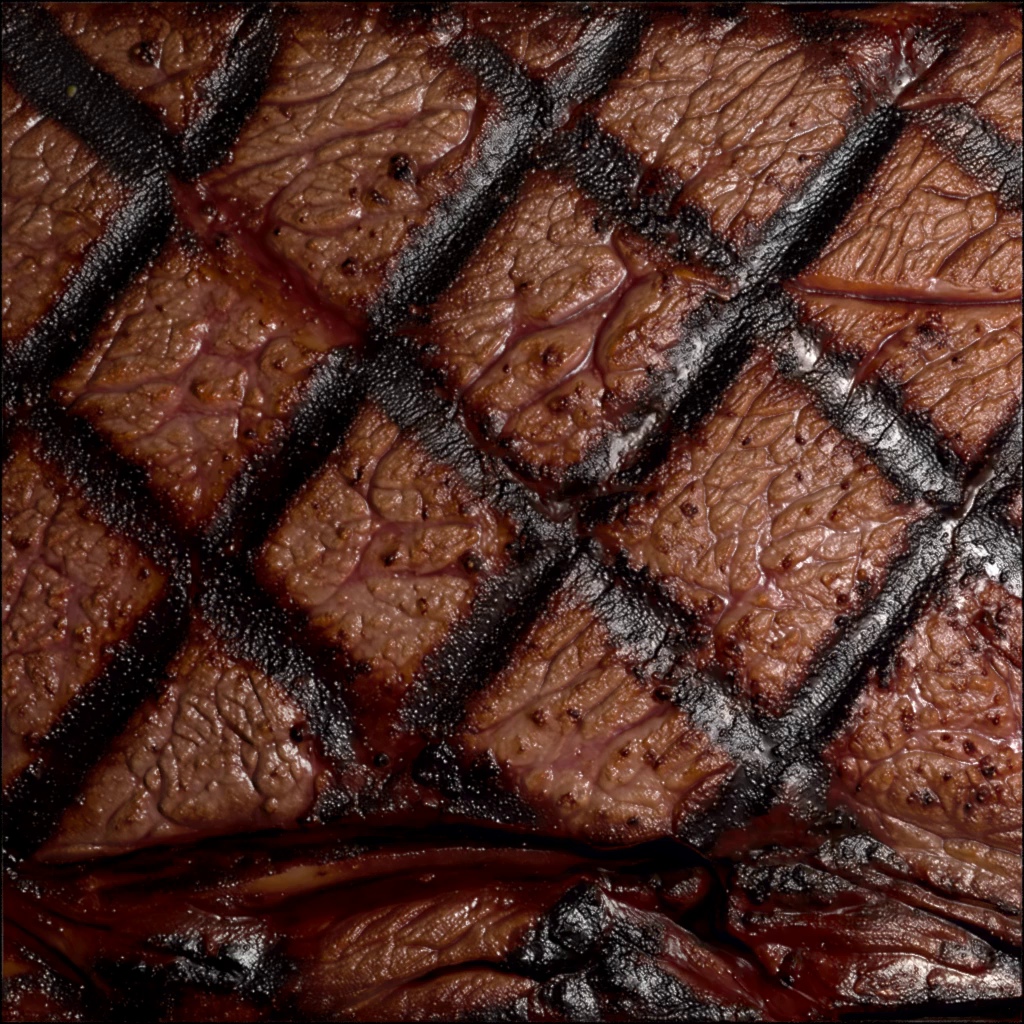} &
        \includegraphics[width=0.100\textwidth]{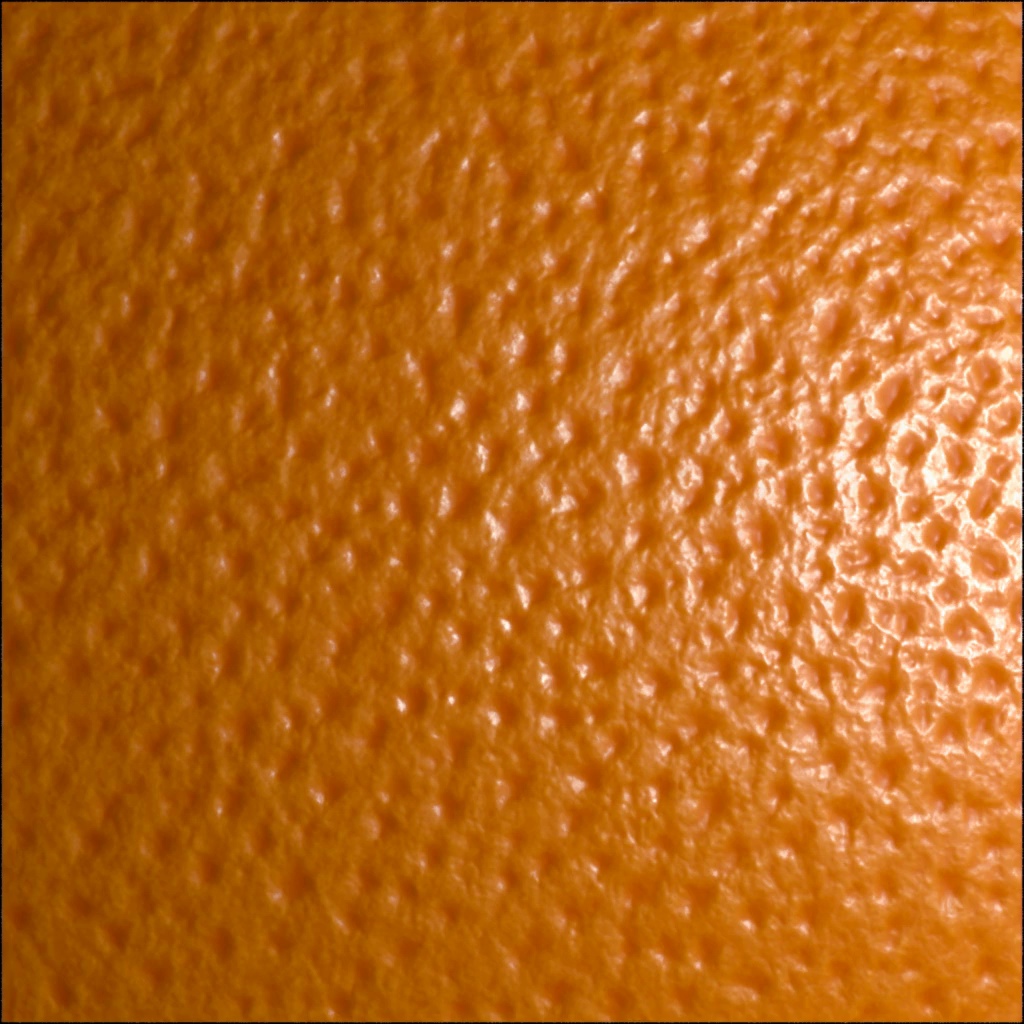} &
        \includegraphics[width=0.100\textwidth]{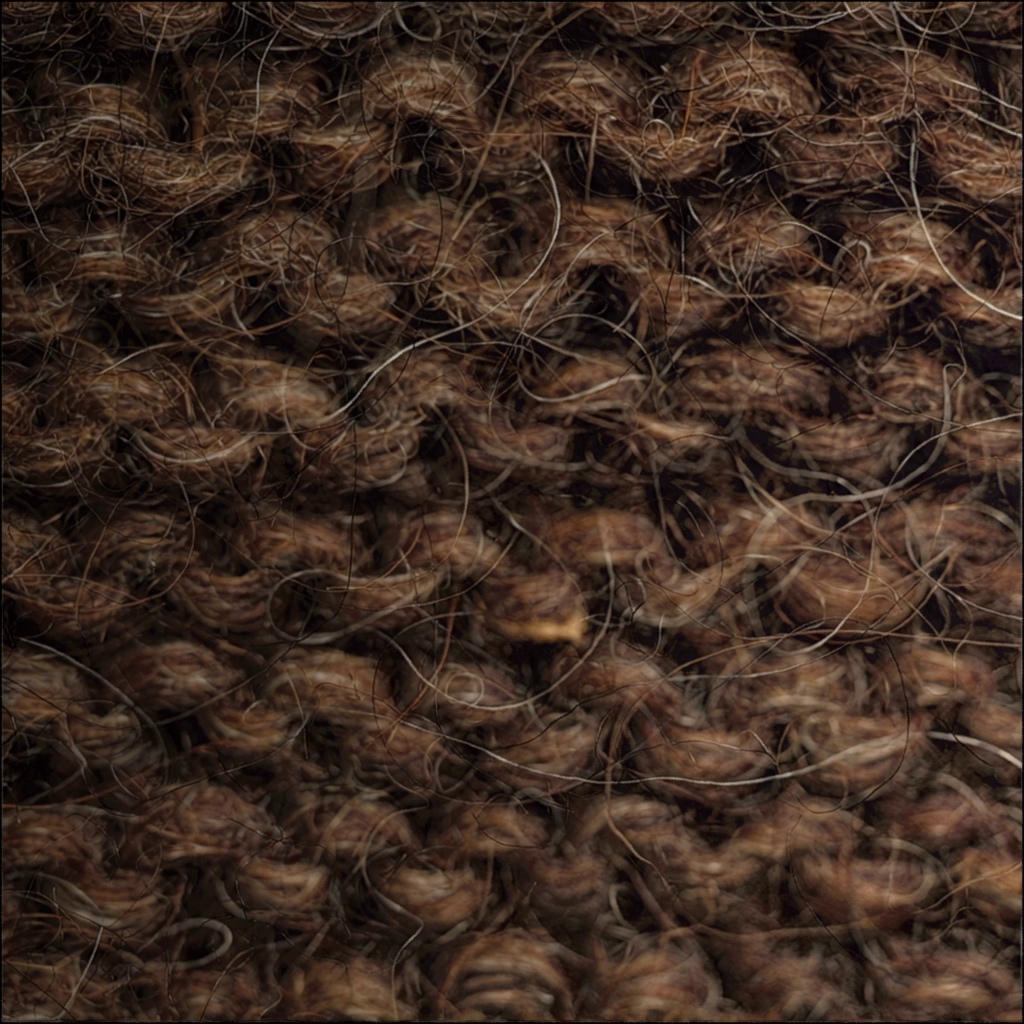} &
        \includegraphics[width=0.100\textwidth]{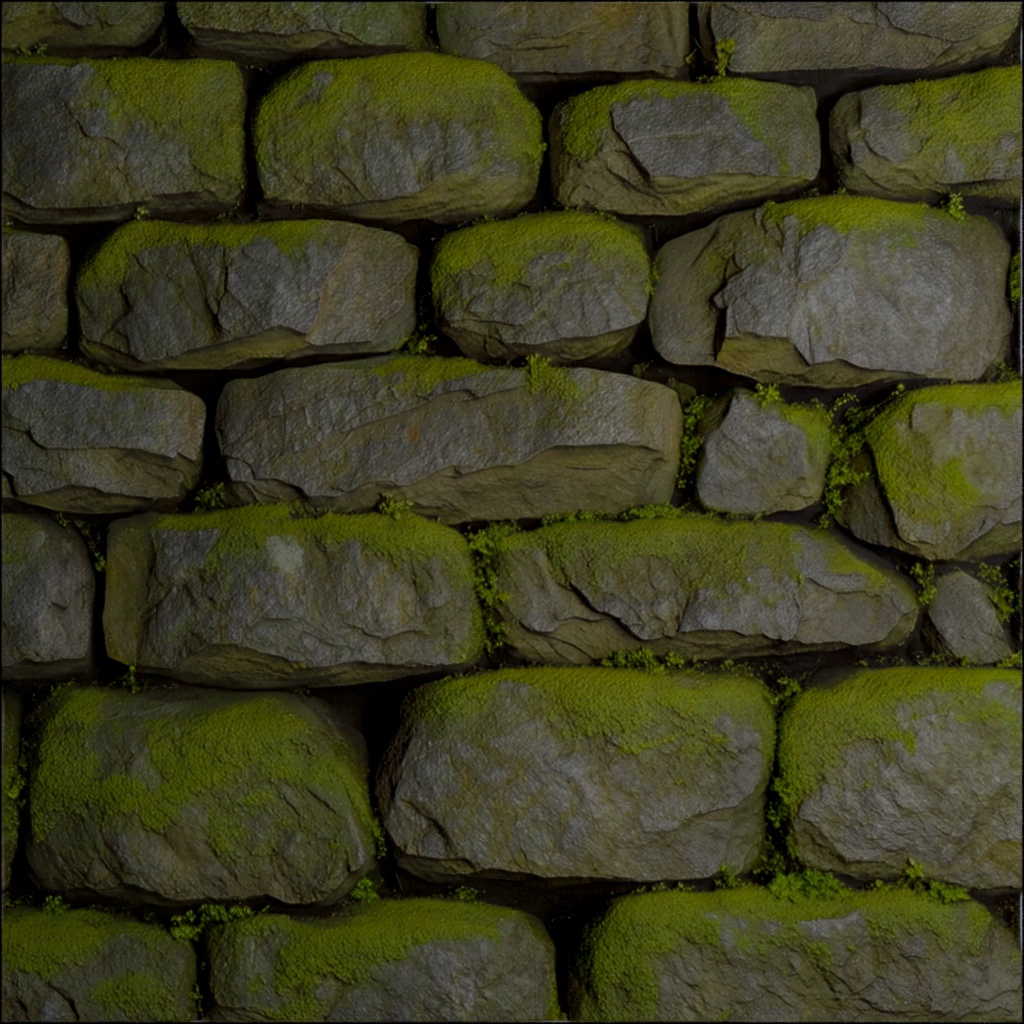} &
        \includegraphics[width=0.100\textwidth]{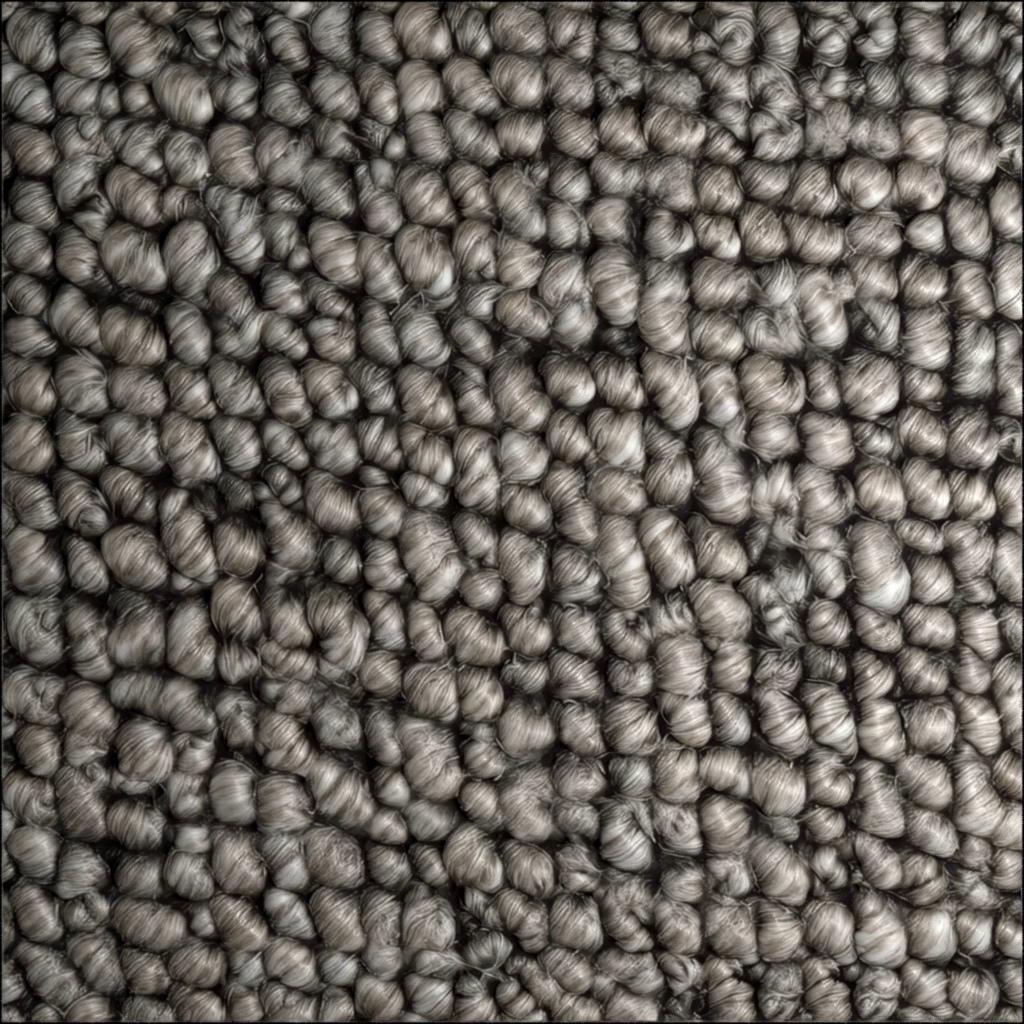} &
        \includegraphics[width=0.100\textwidth]{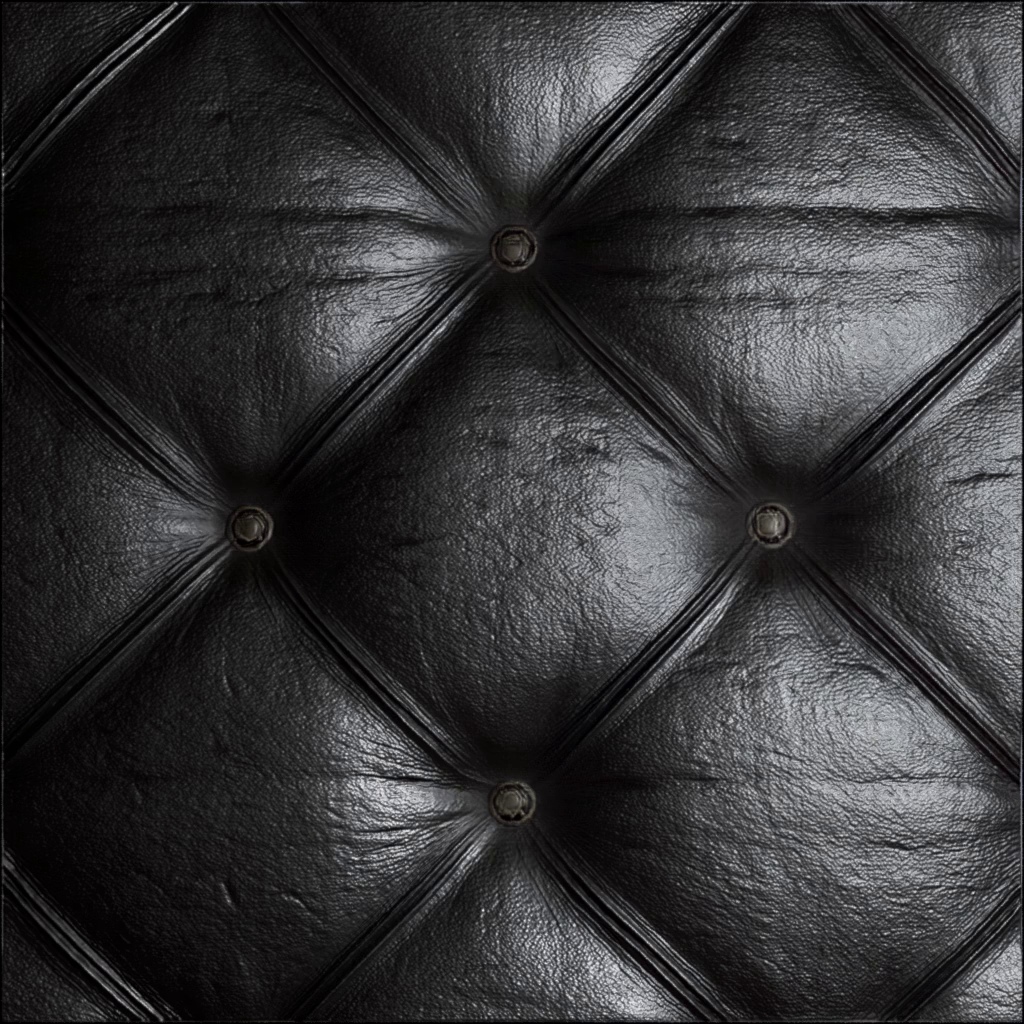} &
        \includegraphics[width=0.100\textwidth]{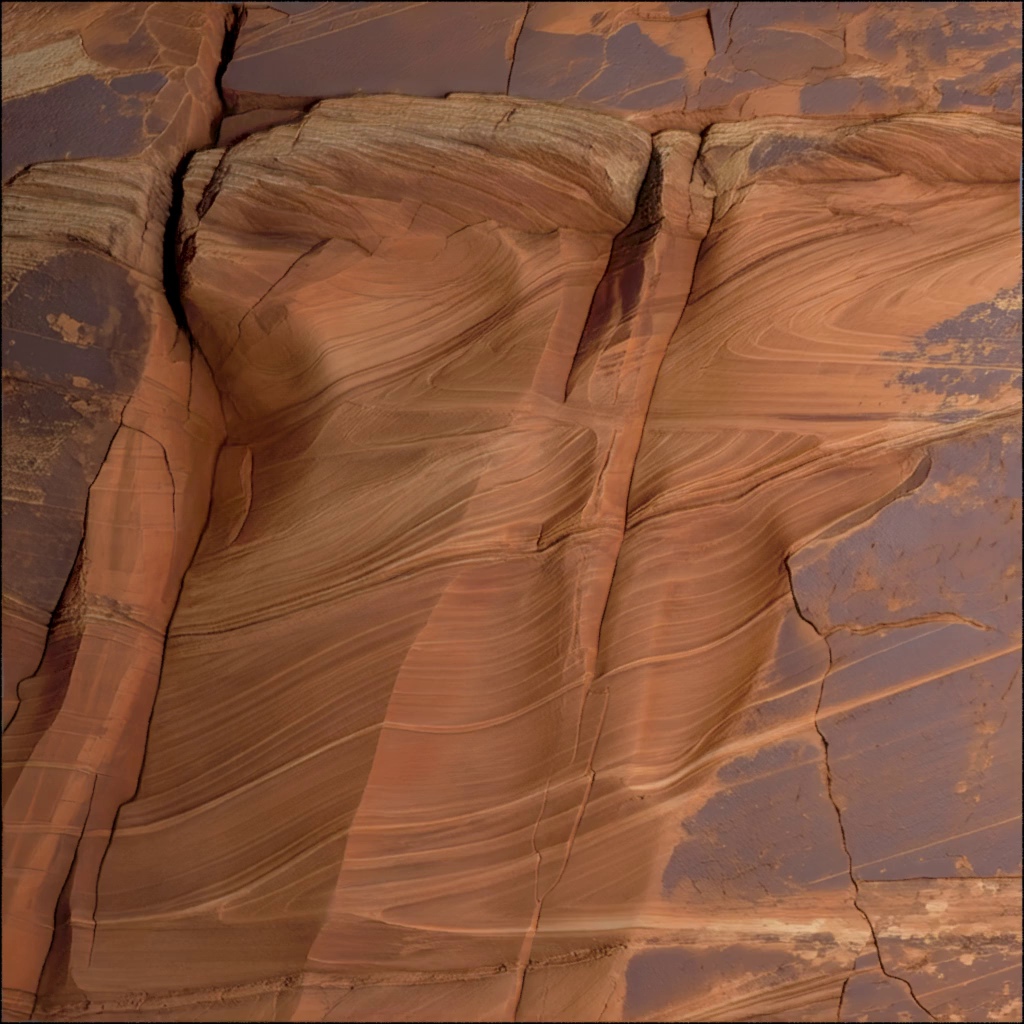} &
        \includegraphics[width=0.100\textwidth]{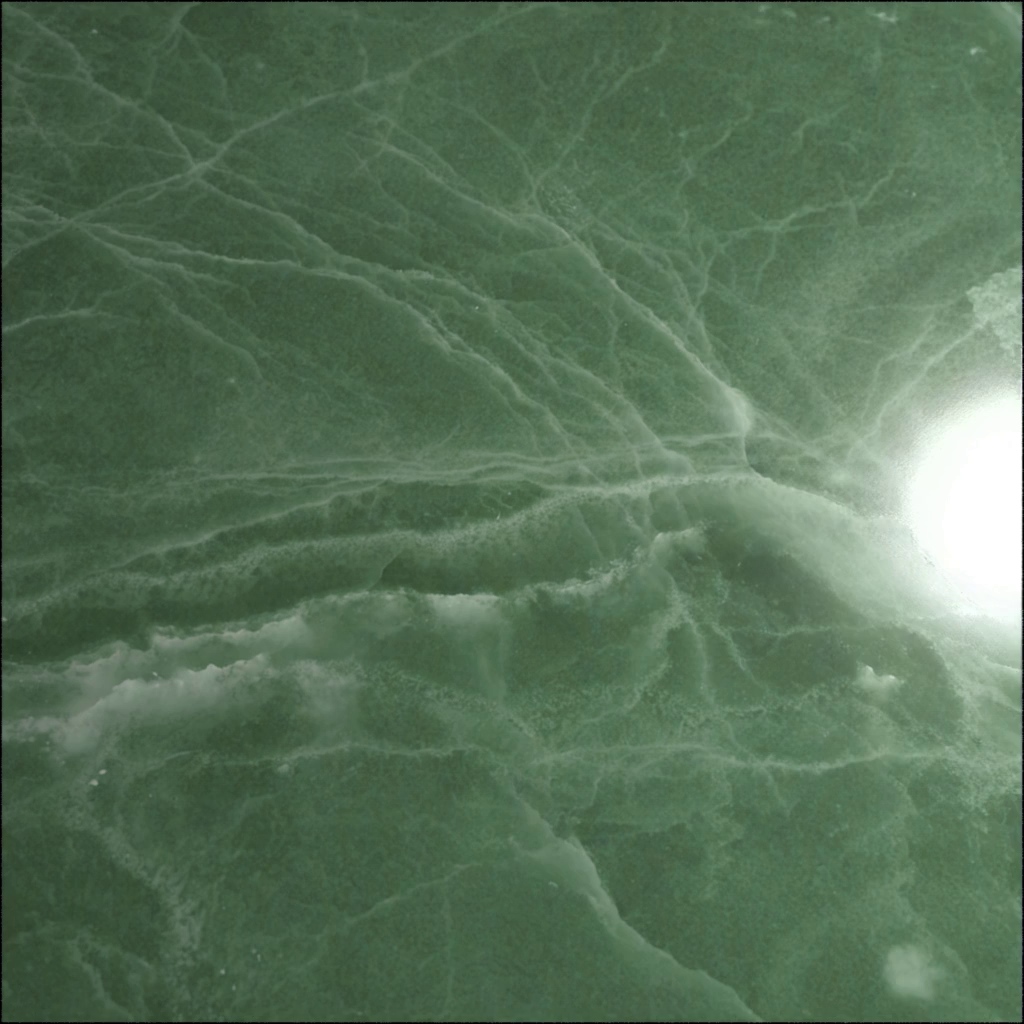} &
        \includegraphics[width=0.100\textwidth]{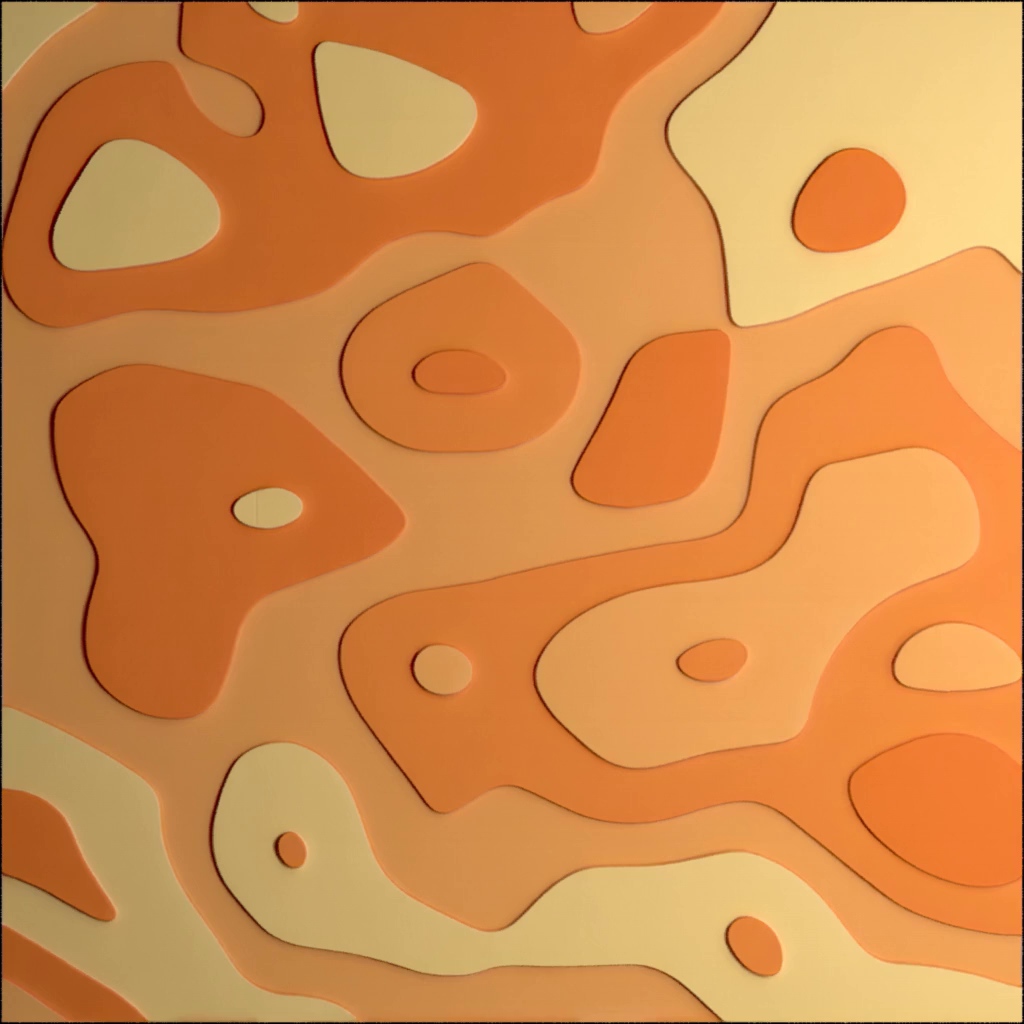} \\    \end{tabular}}

    \caption{\textbf{Image-to-video generation results.} Given a single material photograph as input conditioning (first row), our fine-tuned model generates structured material videos adhering to the learned camera and lighting trajectory. The middle row shows first frame of the video-generated appearance under novel viewing conditions, and the last row shows the LRM-reconstructed neural materials.
    \newline\footnotesize{Image credits: Input condition images 1-8 from left to right are Adobe Stock \cite{adobestock} images, \textcopyright{} BubbleSnap, Earthen Stock, Anna Klimchuk, mr\_studioo, Elena, mila103, Cleopatra and shushan1974 / stock.adobe.com; the last input condition image designed by Magnific, www.magnific.com.}}
    \label{fig:i2v_results}
\end{figure*}


\def\reconTwoNumcols{6}

\ifdefined\reconTwoImgwidth\else\newlength{\reconTwoImgwidth}\fi
\setlength{\reconTwoImgwidth}{\dimexpr(\textwidth / \reconTwoNumcols) - 2pt\relax}

\def\reconTwoBasepath{image/reconstruction/diffusion}

\def\reconTwoTrainframe{0}
\def\reconTwoClothframe{0008}

\def\reconTwoImg#1{%
    \IfFileExists{#1}{%
        \includegraphics[width=\reconTwoImgwidth]{#1}%
    }{%
        \includegraphics[width=\reconTwoImgwidth]{placeholder.jpg}%
    }%
}

\def\reconTwoPlaneImg#1{\reconTwoBasepath/plane/#1/train_views_\reconTwoTrainframe.jpg}
\def\reconTwoClothImg#1{\reconTwoBasepath/cloth/#1/image_\reconTwoClothframe.jpg}

\def\reconTwoColRecon{Reconstruction}
\def\reconTwoColCurved{Curved surf.}

\def\reconTwoMatA{0001_seed42_full_10000iter_cfg5_complex}
\def\reconTwoMatB{0002_seed324_full_10000iter_cfg5_4_1024}
\def\reconTwoMatC{0002_seed759729275_comparision2unet9k}
\def\reconTwoMatD{0003_seed42_full_10000iter_cfg5_vcomplex_res1024}
\def\reconTwoMatE{leather}
\def\reconTwoMatF{0004_seed832623965_v4}
\def\reconTwoMatI{0010_seed269012744_comparisionunet_original}
\def\reconTwoMatJ{0010_seed42_lora_9000iter_res1024_cfg5_vcomplex}
\def\reconTwoMatK{0011_seed581773592_lora_9000iter_res512_cfg7_original_prompt_1024inference}
\def\reconTwoMatN{0018_seed996_comparisionunettileexpanded_prompts_v5}
\def\reconTwoMatO{0023_seed0_lora9000res512cfg5ori1024infmdm}
\def\reconTwoMatP{0024_seed42_full_rectify11000complex1k}
\def\reconTwoMatQ{0025_seed324_lora_9000iter_res1024_cfg5_4}
\def\reconTwoMatR{0027_seed996_comparisionunettileexpanded_prompts_v5}
\def\reconTwoMatS{0036_seed42_lora9000res512cfg5ori1024infcmp}
\def\reconTwoMatT{0031_seed324}
\def\reconTwoMatU{0047_seed324_lora_9000iter_res1024_cfg5_3}
\def\reconTwoMatV{White_fluffy_faux_fur_texture_seed20_i2vour1024fix}

\def\reconTwoPromptA{"... corrugated metal ridges ..."}
\def\reconTwoPromptB{"... green leaves and fruits ..."}
\def\reconTwoPromptC{"... dragon carved wood ..."}
\def\reconTwoPromptD{"... dense animal fur ..."}
\def\reconTwoPromptE{"... button-tufted leather upholstery ..."}
\def\reconTwoPromptF{"... reindeer moss lichen ..."}
\def\reconTwoPromptI{"... stone wall background ..."}
\def\reconTwoPromptJ{"... crimson velvet fabric ..."}
\def\reconTwoPromptK{"... basketweave rattan panel ..."}
\def\reconTwoPromptN{"... frost-touched ivy crystals ..."}
\def\reconTwoPromptO{"... cross-knurled metal ..."}
\def\reconTwoPromptP{"... pitted tufa stone ..."}
\def\reconTwoPromptQ{"... brown woven leather ..."}
\def\reconTwoPromptR{"... chocolate croissant filling ..."}
\def\reconTwoPromptS{"... black cow fur ..."}
\def\reconTwoPromptT{"... pizza crust charred leopard spots ..."}
\def\reconTwoPromptU{"... black white carpet ..."}
\def\reconTwoPromptV{"... white fluffy faux fur ..."}

\def\reconTwoMakerow#1#2#3#4#5#6{%
    \reconTwoImg{\reconTwoPlaneImg{#1}} &
    \reconTwoImg{\reconTwoClothImg{#1}} &
    \reconTwoImg{\reconTwoPlaneImg{#3}} &
    \reconTwoImg{\reconTwoClothImg{#3}} &
    \reconTwoImg{\reconTwoPlaneImg{#5}} &
    \reconTwoImg{\reconTwoClothImg{#5}} \\[-2pt]
    \multicolumn{2}{c}{\scriptsize\textit{#2}} &
    \multicolumn{2}{c}{\scriptsize\textit{#4}} &
    \multicolumn{2}{c}{\scriptsize\textit{#6}} \\[0pt]
}

\begin{figure*}[p]
    \centering
    \setlength{\tabcolsep}{1pt}
    \renewcommand{\arraystretch}{0.85}
    \begin{tabular}{cccccc}
        \small\reconTwoColRecon & \small\reconTwoColCurved &
        \small\reconTwoColRecon & \small\reconTwoColCurved &
        \small\reconTwoColRecon & \small\reconTwoColCurved \\[2pt]
        \hline \\[-6pt]
        \reconTwoMakerow{\reconTwoMatA}{\reconTwoPromptA}{\reconTwoMatB}{\reconTwoPromptB}{\reconTwoMatC}{\reconTwoPromptC}
        \reconTwoMakerow{\reconTwoMatD}{\reconTwoPromptD}{\reconTwoMatE}{\reconTwoPromptE}{\reconTwoMatF}{\reconTwoPromptF}
        \reconTwoMakerow{\reconTwoMatI}{\reconTwoPromptI}{\reconTwoMatJ}{\reconTwoPromptJ}{\reconTwoMatK}{\reconTwoPromptK}
        \reconTwoMakerow{\reconTwoMatN}{\reconTwoPromptN}{\reconTwoMatO}{\reconTwoPromptO}{\reconTwoMatP}{\reconTwoPromptP}
        \reconTwoMakerow{\reconTwoMatQ}{\reconTwoPromptQ}{\reconTwoMatR}{\reconTwoPromptR}{\reconTwoMatS}{\reconTwoPromptS}
        \reconTwoMakerow{\reconTwoMatT}{\reconTwoPromptT}{\reconTwoMatU}{\reconTwoPromptU}{\reconTwoMatV}{\reconTwoPromptV}
    \end{tabular}
    \caption{\textbf{Reconstruction results.} A selection of materials generated by our pipeline from text prompts, shown on a flat and curved surface under different illuminations. Note the realism of the results and the ability to handle non-trivial geometry (leaves, fur, fabric) that cannot be represented by heightfields. Please see the extensive supplementary materials for animated results, showing parallax effects.} 
    \label{fig:recon_results_2}
\end{figure*}



\paragraph{LRM ablations (\autoref{tab:ablation}).}
 The table shows that not using pretrained weights for either NeuMIP MLP or Wan DiT/VAE decoder has the lowest scores, as explained in \autoref{sec:lrm}. The pretrained MLP was trained on 512 randomly selected MatSynth materials. It also reveals freezing the pre-trained MLP during LRM training leads to degraded results. Additionally, utilizing a smaller/larger NeuMIP MLP, or using fewer or more input views do not change the metrics dramatically. Lastly, using a linear layer to upsample the DiT tokens instead of the VAE decoder leads to lower scores, except for LPIPS. Comparing renderings of either upsamplers in \autoref{fig:upsampling} reveals the patchy artifacts of linear upsampler as each upsampled token is processed independently of others.

\section{Limitations and Future Work}
The NeuMIP representation has limits in its ability to approximate true 3D effects; a more sophisticated neural representation can be designed, as our LRM design is representation-agnostic. Our pipeline is finetuned solely on MatSynth, which lacks complexity such as translucent, layered or fiber materials. While the video diffusion model can still generate these complex materials due to its generative prior, augmenting MatSynth, or training the LRM to predict generated video frames from other generated frames could further improve the complexity of the generated materials. Lastly, further improvements on our tileability solution may be possible.
 
\section{Conclusion}
We presented VideoNeuMat, a pipeline that extracts reusable neural materials from video diffusion models. By finetuning a pretrained video model as a "virtual gonioreflectometer" and introducing a large reconstruction model for single-pass neural material inference from such generated videos, we successfully transfer material knowledge from internet-scale video data into standalone material assets, usable on new surfaces under new views and lighting. Our results demonstrate higher realism than synthetic training data and previous diffusion-based solution, breaking through the data scarcity barrier inherent to the material generation problem.

\begin{acks}
We would also like to thank Aaron Lefohn for his support, and NVIDIA for supporting the work through an NVIDIA academic partnership.    
\end{acks}



\bibliographystyle{ACM-Reference-Format}
\bibliography{main}

\end{document}